\documentclass[letterpaper]{article} 
\usepackage{aaai25}  
\usepackage{times}  
\usepackage{helvet}  
\usepackage{courier}  
\usepackage[hyphens]{url}  
\usepackage{graphicx} 
\usepackage{natbib}  
\usepackage{caption} 
\usepackage{algorithm}
\usepackage{algorithmic}

\usepackage{newfloat}
\usepackage{listings}
\DeclareCaptionStyle{ruled}{labelfont=normalfont,labelsep=colon,strut=off} 
\floatstyle{ruled}
\newfloat{listing}{tb}{lst}{}
\floatname{listing}{Listing}
\usepackage[utf8]{inputenc} 
\usepackage[T1]{fontenc}    
\usepackage{url}            
\usepackage{amsfonts}       
\usepackage{nicefrac}       
\usepackage{microtype}      
\usepackage{xcolor}         
\usepackage{pgf}

\usepackage{multirow}
\usepackage{dsfont}
\usepackage{amsmath, amsthm, mathtools, amssymb}
\usepackage{svg}
\usepackage{amsfonts}
\usepackage[shortlabels, inline]{enumitem}
\usepackage{tikz}
\usepackage{collcell}
\usepackage{graphicx}
\usepackage{caption}
\usepackage{listings}
\usepackage[export]{adjustbox}
\usepackage{colortbl}
\usepackage{scalerel}

\definecolor{dkgreen}{rgb}{0,0.6,0}
\definecolor{gray}{rgb}{0.5,0.5,0.5}
\definecolor{mauve}{rgb}{0.58,0,0.82}
\definecolor{lightgray}{gray}{0.9}
\definecolor{BrickRed}{HTML}{E45756} 
\definecolor{ForestGreen}{HTML}{54A24B} 
\definecolor{Mulberry}{HTML}{4C78A8} 
\definecolor{WildStrawberry}{HTML}{9467BD}
\definecolor{Cerulean}{HTML}{B279A2}
\definecolor{MidnightBlue}{HTML}{B279A2}

\definecolor{GoodBlue}{HTML}{3772FF} 
\definecolor{BadRed}{HTML}{DF2935} 

\definecolor{LightGrayCell}{HTML}{D3D3D3} 
\definecolor{MiddleCell}{HTML}{E4E4E4} 
\definecolor{WhiteSmokeCell}{HTML}{F5F5F5} 

\newcommand{\gradient}[1]{%
  \pgfmathsetmacro{\sign}{sign(#1)}%
  \pgfmathsetmacro{\absvalue}{abs(#1)}%
  \pgfmathsetmacro{\col}{100*(\absvalue-(-10))/(10-(-10))}%
  \ifdim \sign pt < 0 pt %
    \ifdim \absvalue pt > 9 pt %
      \textcolor{BadRed}{\textbf{#1}}%
    \else
      \textcolor{BadRed!\col!GoodBlue}{\textbf{#1}}%
    \fi
  \else
    \ifdim \absvalue pt > 3 pt %
      \textcolor{GoodBlue}{\textbf{#1}}%
    \else
      \textcolor{GoodBlue!\col!BadRed}{\textbf{#1}}%
    \fi
  \fi
}

\newcommand{\gradientcolor}[1]{%
  \pgfmathsetmacro{\col}{(#1)}%
  \textcolor{BadRed!\col!GoodBlue}{\textbf{#1}}%
}

\title{Is There No Such Thing As A Bad Question?\\H4R: HalluciBot For Ratiocination, Rewriting, Ranking, and Routing}
\author{
    William Watson\equalcontrib,
    Nicole Cho\equalcontrib,
    Nishan Srishankar
}
\affiliations{
    J.P. Morgan AI Research\\

    nicole.cho@jpmorgan.com
}

\usepackage{bibentry}

\begin{document}

\maketitle

\begin{abstract}
Hallucination continues to be one of the most critical challenges in the institutional adoption journey of Large Language Models (LLMs). While prior studies have primarily focused on the post-generation analysis and refinement of outputs, this paper centers on the effectiveness of queries in eliciting accurate responses from LLMs. We present \textbf{HalluciBot}, a model that estimates the query's propensity to hallucinate \textbf{\textit{before generation}}, without invoking any LLMs during inference. HalluciBot can serve as a proxy reward model for query rewriting, offering a general framework to estimate query quality based on accuracy and consensus.  
In essence, HalluciBot investigates how poorly constructed queries can lead to erroneous outputs - moreover, by employing query rewriting guided by HalluciBot's empirical estimates, we demonstrate that $95.7\%$ output accuracy can be achieved for \textbf{\color{ForestGreen}Multiple Choice} questions. The training procedure for HalluciBot consists of perturbing 369,837 queries $n$ times, employing $n+1$ independent LLM agents, sampling an output from each query, conducting a Multi-Agent Monte Carlo simulation on the sampled outputs, and  training an encoder classifier. The idea of perturbation is the outcome of our ablation studies that measures the increase in output diversity ($+12.5$ agreement spread) by perturbing a query in lexically different but semantically similar ways.
Therefore, HalluciBot paves the way to ratiocinate ($76.0\%$ test F1 score, $46.6\%$ in saved computation on hallucinatory queries), rewrite ($+30.2\%$ positive class transition from hallucinatory to non-hallucinatory), rank ($+50.6\%$ positive class transition from hallucinatory to non-hallucinatory), and route queries to effective pipelines.
\end{abstract}

%
\begin{links}
    \link{Extended version}{https://arxiv.org/abs/2404.12535}
\end{links}

\section{Introduction}
\label{sec:intro}

\begin{figure}[!t]
\centering
\includegraphics[clip, trim=9.95cm 5.19cm 6.34cm 3.095cm,width=\linewidth]{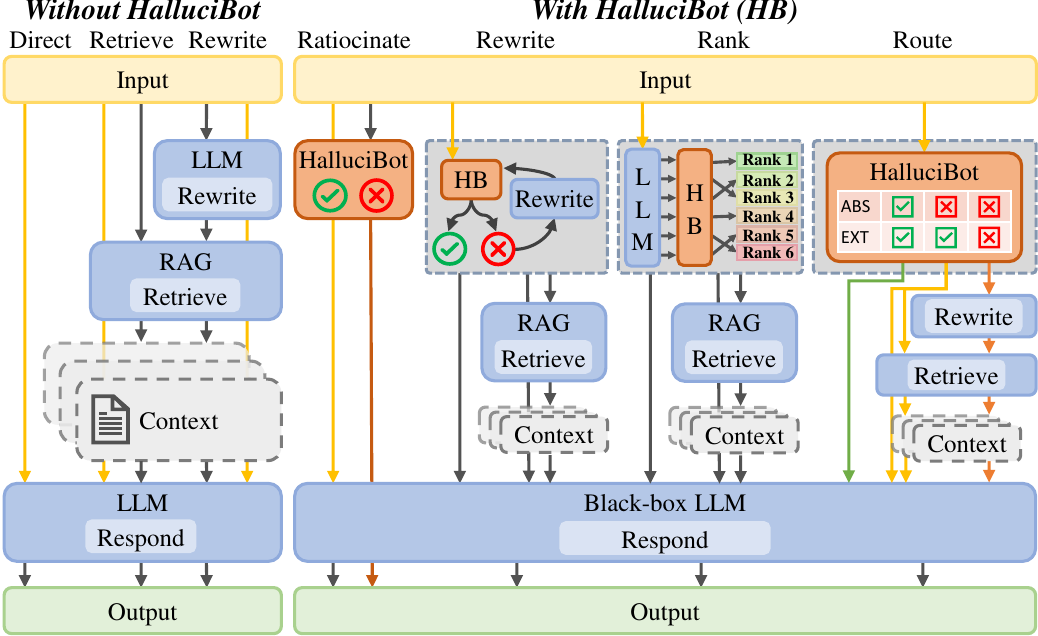}
\caption{
Comparison of traditional inference methods and HalluciBot's use-cases. 
In the former, the user inputs a query either through a direct inference or a retrieval-augmented generation (RAG) pipeline.  
If the output is hallucinatory, the user must decide whether to end the session or revise the query for successive generation rounds.
In contrast, HalluciBot can be used to assess the query's quality \textbf{\textit{before generation}}. 
Therefore, users can gain insight into the hallucination risk (``Ratiocinate''), automate the query rewriting stage  through informed feedback (``Rewrite'') or Best-of-N sampling across multiple candidates (``Rank''), and route the query across different operating modes (``Route''), since HalluciBot is scenario-aware (\textbf{\color{BrickRed}\underline{Ext}ractive} / \textbf{\color{Mulberry}\underline{Abs}tractive}), potentially bypassing computationally expensive stages, such as RAG or Rewrite. 
} 
\label{fig:compare}
\vspace{-3mm}
\end{figure}

Despite the promising potential for a myriad of use cases, Large Language Models (LLMs) offer limited insights into their chain of thought \citep{liang2022holistic, wei2023chainofthought, kojima2023large, li2023making} and have the propensity to hallucinate in various circumstances \citep{10.1162/tacl_a_00407}. Common factors that drive hallucinations encompass high model complexity, flawed data sources, or inherent sampling randomness. Specifically, the intrinsic trade-off between greedy deterministic decoding and the creativity spawned through nucleus sampling induces a heightened propensity to hallucinate \citep{huang2023survey} - LLMs frequently advance output quality through different sampling methods \citep{holtzman2020curious,fan-etal-2018-hierarchical,holtzman-etal-2018-learning,radford-topk}. The challenge of understanding hallucinations is compounded by limitations such as the frequent inaccessibility into the LLMs' training datasets \citep{liang2022holistic}. HuggingFace's release of its ``Hallucinations Leaderboard'' on January 29th, 2024 \citep{hallucinations-leaderboard,eval-harness} highlights the importance of resolving hallucination-related issues via the concerted effort of evaluating different LLMs. In this context, the majority of current studies have focused on the post-generation phase of output analysis as expanded in \citet{peng2023check} such as - (1) self-refinement via feedback loops on the model's output \citep{madaan2023selfrefine}, (2) analysis of logit output values to detect hallucination \citep{varshney2023stitch}, or (3) for a minority of studies focused on the pre-generation phase, the ingestion of recent knowledge to improve performance \citep{tonmoy2024comprehensive}. We propose a novel model, \textbf{HalluciBot}, that predicts the probability of hallucination, \textit{\textbf{before any generation}}, for a given query.
In essence, this paper refocuses the study of hallucination to an empirical evaluation of the input query - how much does the query's quality influence the model's propensity to hallucinate? Therefore, HalluciBot estimates, 
\begin{itemize}[noitemsep, leftmargin=*, topsep=0pt, partopsep=0pt, label={\tiny\raisebox{0.5ex}{$\blacktriangleright$}}]
\item a binary classification of the query's propensity to hallucinate (``Yes'' or ``No''), as well as,
\item a non reinforcement-learning method to guide query rewriting, enabling the construction of this encoder to be agnostic to closed-source or open-source LLMs \citep{ma2023query}.
\end{itemize}
We train HalluciBot as a binary classifier to predict whether a query will lead to erroneous outputs. To generate ground truth labels, we use a Multi-Agent Monte Carlo simulation that perturbs the query and checks for inaccuracies. If any perturbed version causes an error, the original query is labeled as hallucinatory. In this paper, HalluciBot leverages \texttt{gpt-3.5-turbo}, trained via (1) perturbing 369,837 queries $n$ times to retain the original semantic meaning yet diverge lexically, (2)
employing $n+1$ independent agents to sample an output from each perturbation including the original query, at a temperature of $1.0$ for diversity, (3)  conducting a Monte Carlo simulation on 2,219,022 sampled outputs, and (4) deriving an empirical estimate into the expected rate of hallucination $p_h(q_0)$ for the original query. 
We prove that introducing perturbations before sampling $n + 1$ outputs for query $q_0$ garners a $13.2$ point spread in the lower and upper bound accuracy, with a $12.5$ point decrease in Fleiss’s $\kappa$ for agreement, even as the modal accuracy remains largely unchanged ($1.3$ point difference).
In other words, perturbations introduce more variability in the outputs, while preserving the central tendency.
As HalluciBot generates the probability of hallucination in the pre-generation phase, the estimates can be used in a myriad of downstream modalities (Figure~\ref{fig:compare}) such as:
\textbf{``Ratiocinate''} to purely estimate the query's quality;
\textbf{``Rewrite''} to leverage the probabilities and improve the query's quality via iterative feedback;
\textbf{``Rank''} to rank perturbations, using probabilities as a proxy reward model in \textit{Best-of-N} sampling; 
\textbf{``Route''} to route the best next steps, depending on scenarios such as \textbf{\textcolor{BrickRed}{Extractive}} or \textbf{\textcolor{Mulberry}{Abstractive}}. By cross-tabulating the predicted hallucination rates across scenarios, HalluciBot can act as a router, through which certain queries can be guided to a black-box LLM, while others will require a more complex pipeline including context retrieval, web search, or agents \citep{watson-etal-2023-hiddentables, zeng2024flowmind, Cho_2024}.

\textbf{Contributions.} As a result, our study has culminated in the following  pillars of contribution.  
\textbf{(1)} HalluciBot is the first encoder-based model to derive, \textit{\textbf{before generation}}, an anticipated rate of hallucination \textit{for any type of query}, achieving a validation accuracy of $73.6\%$ ($80.2\%$ F1) and a testing accuracy of $69.5\%$ ($76.0\%$ F1). \textbf{(2)} our approach to construct HalluciBot absorbs the computational complexity of Monte Carlo sampling, exploration, and training prior to the user session. Thus, institutions that employ HalluciBot can systematically save on the considerable amount of computational waste engendered by ``highly probable'' hallucinatory queries ($46.6\%$ in saved computation during testing). 
\textbf{(3)} the hallucination probability can be leveraged as a proxy reward model in a myriad of different infrastructures; HalluciBot
paves the way to 
rewrite ($+31.9\%$ positive class transition from hallucinatory to non-hallucinatory), rank ($+51.4\%$ positive class transition from hallucinatory to non-hallucinatory), and route ($+60.0\%$ 
diverted) queries. 
\textbf{(4)} HalluciBot generalizes to systems with RAG or few-shot question answering systems with an LLM generator by differentiating the scenario in its prompt. 
Also, it can generalize to closed systems only accessible via API calls \citep{openai,google,microsoft}. 
\textbf{(5)} HalluciBot's training methodology can be leveraged for any model or training corpus; it can be leveraged as a general means by which the research community can develop an encoder to assess query quality.

\begin{figure}[t]
\centering
\includegraphics[clip, trim=9.75cm 7.31cm 4.02cm 4.15cm,width=\linewidth]{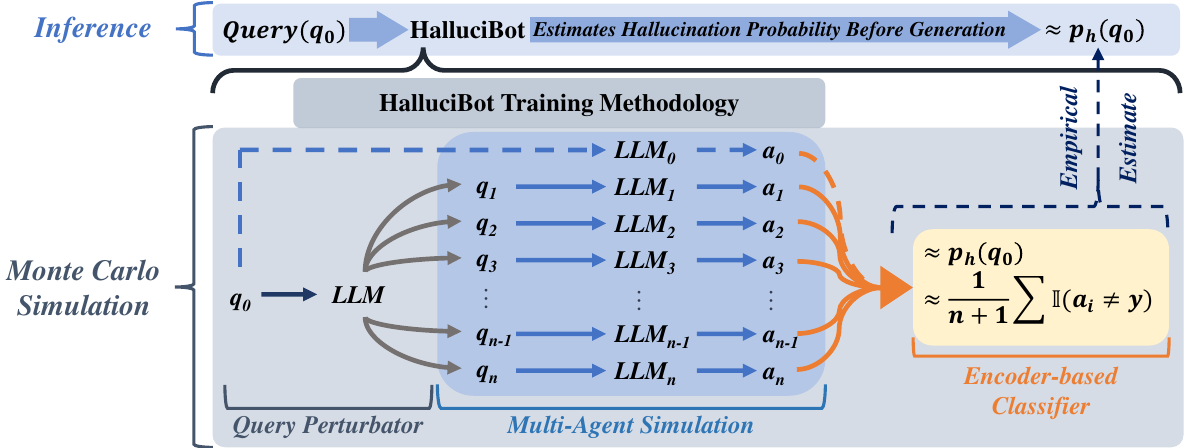}
\vspace{-5mm}
\caption{Training Overview. A single query, $q_0$, is \textbf{perturbed} in $n$ different ways. Next, The original and perturbed queries $q_i$ are independently answered by the Generator agents. This Multi-Agent Monte Carlo simulation provides an estimate into the \textbf{rate of hallucination} $p_h(q_0)$ for the original query $q_0$. Via these simulated results, 
HalluciBot is trained to predict the probability that any $q_0$ \textit{could} hallucinate, and predict the expected consensus of sampled outputs \textit{\textbf{before generation}}.
}
\label{fig:system_overview}
\end{figure}

\section{Related Work}

With regards to hallucination mitigation studies, an overwhelming majority focuses on the post-generation stage of analyzing outputs. A minority concentrates on the pre-generation phase and even amongst those, the focus lies in incorporating recent knowledge into LLMs. In detail, many expand on the universally utilized method of context-based retrieval systems \citep{reimers-2019-sentence-bert, johnson2019billion, nogueira2020passage, karpukhin-etal-2020-dense, NEURIPS2020_6b493230, izacard-grave-2021-leveraging}. 
Other methods include relying on the model's general knowledge \citep{khashabi-etal-2020-unifiedqa} or conditioning the QA model on context generated by the LLM itself \citep{yu2023generate}. Certain work has focused on mitigating hallucinations by augmenting the way LLMs generate their answers. One of the more popular techniques is to have the model enumerate its chain-of-thought \citep{wei2023chainofthought} and think step by step \citep{nye2021work}, while building context. Another method to augment generation with context is by semantic retrieval \citep{NEURIPS2020_6b493230,liu2021makes}, handling hallucinations as they arise \citep{varshney2023stitch}, or using LLMs to generate context rather than retrieve \citep{yu2023generate}. PromptChainer \citep{10.1145/3491101.3519729} profiled techniques to craft LLM chains, in which the output of one LLM's generation process, when fed into the next LLM, can allow for more complex tasks. Language Model Cascades \citep{dohan2022language} demonstrated that LLMs can yield probabilistic programs to tackle multi-step reasoning problems. Self-consistency \citep{wang2023selfconsistency} leveraged a new decoding strategy to sample multiple generative pathways - then select the most consistent answer. Also, \citet{kumar-etal-2022-gradient} explored gradient-based sampling procedures that satisfy user-defined constraints.
Most recent work has focused on sampling-based calibration within a single model \citep{cole-etal-2023-selectively} or self-verification \citep{kadavath2022language} - the latter focuses on generating a set of outputs and feeding those back into the LLM. Furthermore, \citet{snyder2023early} explores how artifacts can differentiate hallucinated outputs. 
One common feature amongst these approaches is that the focus is on the output rather than the query. \citet{alzahrani2024benchmarks} explored how LLMs are highly sensitive to minute perturbations, such as changing the order of answer choices. Also, while \citet{zheng2023noisy} study lexical perturbations, no study on hallucinations employs a Multi-Agent approach coupled with query perturbations - which are hallmark features of HalluciBot. 

\begin{table}[t]
    \centering
    \small
    \begin{tabular}{l|l}
    \hline
    \multicolumn{1}{c|}{\textbf{Scenario}} &  \multicolumn{1}{c}{\textbf{Datasets}} \\
    \hline
    {\color{BrickRed} \textbf{Extractive}} &  \textit{SQuADv2}  \\
    \hline
    {\color{ForestGreen} \textbf{Multiple Choice}} 
    &  \textit{TruthfulQA}, \textit{SciQ}, MMLU, PIQA, 
     BoolQ, \\
     & OpenBookQA, MathQA,
     ARC - E/C\\
    \hline
    {\color{Mulberry} \textbf{Abstractive}} 
    & \textit{SQuADv2}, \textit{TruthfulQA}, \textit{SciQ}, \\
    & 
     WikiQA, HotpotQA, TriviaQA \\
    \hline 
    \end{tabular}
    \vspace{-3pt}
    \captionof{table}{\label{tab:quick-ds}
    Dataset Scenario Split with \textit{Reused Assets}.
    }
    \vspace{-2mm}
\end{table}

\begin{figure*}[!t]
    \centering
        \centering
        \includegraphics[clip, trim={0.55cm 0.51cm 2.1cm 1.05cm}, height=5.4cm]{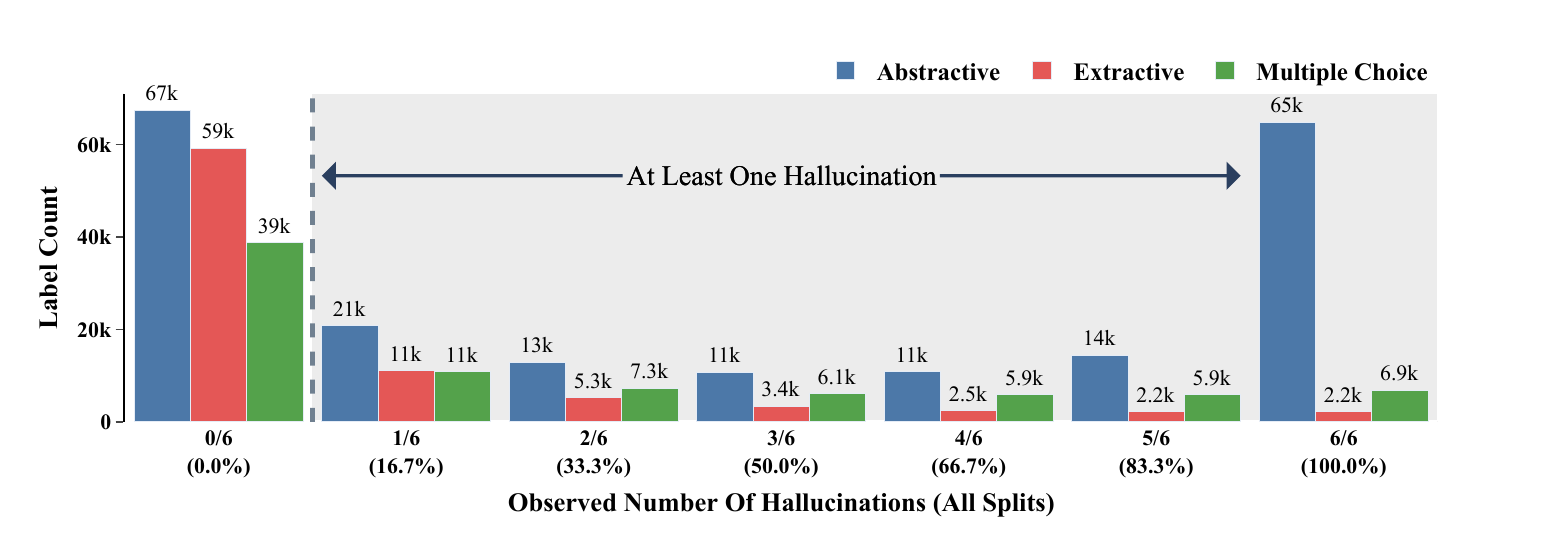}
        \vspace{-1mm}
        \caption{Distribution of the observed number of hallucinations per scenario. For {\color{BrickRed}\textbf{Extractive}}, additional context mitigates the rate of hallucination. For {\color{ForestGreen}\textbf{Multiple Choice}}, distractors can cause confusion amongst agents uniformly. However, for {\color{Mulberry}\textbf{Abstractive}}, no additional information can cause massive disparities in correctness - most of our simulations resulted in no or all hallucinations.}
        \label{fig:figure-multi}
    \vspace{1.25mm}

    \begin{minipage}[b]{.47\linewidth}
        \centering
        \includegraphics[clip, trim={0.5cm 0.97cm 1.9cm 1.36cm}, width=0.95\columnwidth]{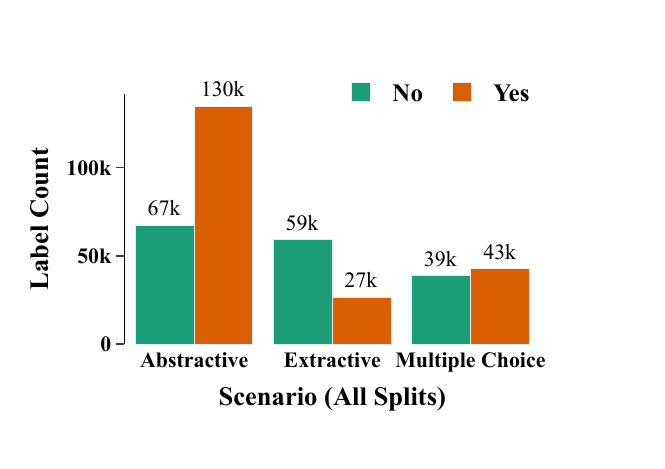}
        \captionof{figure}{Binary distribution of labels, where at least one hallucination occurred during our simulation.}
        \label{fig:figure-binary}
    \end{minipage}\hfill
    \begin{minipage}[b]{.47\linewidth}
        \centering
        \small
        \begin{tabular}{l|rrr}
        \hline
        \textbf{Binary} & \multicolumn{1}{c}{\textbf{Train}} & \multicolumn{1}{c}{\textbf{Val}} & \multicolumn{1}{c}{\textbf{Test}} \\
        \hline
        No $(y=0)$ & 139,142 & 17,153 & 9,306 \\ 
        Yes $(y=1)$ & 163,350 & 27,338 & 13,548 \\
        \hline
        \textbf{Observed Rate} & \multicolumn{1}{c}{\textbf{Train}} & \multicolumn{1}{c}{\textbf{Val}} & \multicolumn{1}{c}{\textbf{Test}} \\
        \hline
        0.0\% $(y=0/6)$   & 139,123 & 17,146  & 9,202   \\
        16.7\% $(y=1/6)$  & 35,114  & 4,974   & 2,757   \\
        33.3\% $(y=2/6)$  & 20,213  & 3,371   & 1,967   \\
        50.0\% $(y=3/6)$  & 15,749  & 2,757   & 1,768   \\
        66.7\% $(y=4/6)$  & 14,477  & 2,735   & 1,970   \\
        83.3\% $(y=5/6)$  & 17,123  & 3,242   & 2,171   \\
        100.0\% $(y=6/6)$ & 60,693  & 10,266  & 3,019   \\
        \hline 
        \textbf{Scenario} & \multicolumn{1}{c}{\textbf{Train}} & \multicolumn{1}{c}{\textbf{Val}} & \multicolumn{1}{c}{\textbf{Test}} \\
        \hline
        {\color{BrickRed} \textbf{Extractive}} & 80,049 & 5,843 & - \\
        {\color{ForestGreen} \textbf{Multiple Choice}} & 45,997 & 14,127 & 21,573 \\
        {\color{Mulberry} \textbf{Abstractive}} & 176,446 & 24,521 & 1,281 \\
        \hline
        Total & 302,492 & 44,491 & 22,854 \\
        \hline 
        \end{tabular}
        \captionof{table}{Training Splits for HalluciBot.}
        \label{tab:splits}
    \end{minipage}

    
    \vspace{-2mm}
\end{figure*}


\section{Methodology Overview}
\label{sec:method}

\textbf{What is Hallucination?}
In general terms, hallucination refers to a false perception of patterns or objects resulting from one's senses. With regards to LLMs, a myriad of studies
bifurcate into (1) \textit{factuality hallucinations} that refer to outputs which directly contradict or fabricate the ground truth while (2) \textit{faithfulness hallucinations} define outputs that misunderstand the context or intent of the query \citep{huang2023survey, Ji_2023}. 
In this study, we introduce \textit{truthful hallucination} as the motivation on why we are perturbing the original query. \textit{Truthful hallucination} is defined as an LLM's inability to answer semantically similar but lexically different perturbations of a query. The motivation for \textit{truthful hallucination} stems from the analysis that neural networks display an intrinsic propensity to memorize training data \citep{carlini2021extracting} - in this case, memorizing the query and output. 
Given the risk of over-training LLMs, their opaque training data, and propensity to memorize - generating multiple outputs from the same query or analyzing a single output from a single query do not help measure \textit{truthful hallucination}. 

\textbf{What is the Motivation for HalluciBot?} \label{sec:hallucibot}
HalluciBot focuses on distilling \textit{LLM behavior} into a speedy encoder that can predict hallucination \textbf{\textit{before generation}}. 
Foremost, this is in contrast to prior work that uses multiple generations during a user's session to provide self-consistency \citep{manakul2023selfcheckgpt}. Next, our proposal differs from entropy based, log-prob based, or model based estimation techniques \citep{huang2023survey} that rely on the LLM's uncertainty to predict hallucinations - these methods focus on the model's bias while we focus on empirical estimates. Moreover, our approach consists of a Multi-Agent simulation which stands in stark contrast to the majority of current experiments that have focused on leveraging a single LLM agent to generate outputs from a single query  \citep{cole-etal-2023-selectively, kadavath2022language, snyder2023early}.
The training procedure for HalluciBot consists of perturbing each query $n=5$ times, employing $n+1=6$ independent LLM agents, sampling an output from each query, conducting a Monte Carlo simulation on 2,219,022 sampled outputs, and  training an encoder classifier.

\subsection{Multi-Agent Monte Carlo Simulation} 
\textbf{What is the Purpose of a Monte Carlo Simulation?}
\label{sec:montecarlo}
As evidenced by multiple studies and Table \ref{tab:aggregate_experiments}, hallucination is the outcome of multiple confounding variables - thus, it is highly unlikely that a tractable closed-form solution will be able to model hallucinations. Thus, we employ a \textbf{Monte Carlo} simulation as a means to derive empirical estimations of hallucination rates in LLMs, since this method is frequently leveraged to map probability in the presence of random variable inference \citep{swaminathan2021monte}. 
Thus, we estimate the probability density that a query induces hallucination.

\textbf{What is a Query Perturbator?}
\label{sec:queryperturbator}
Via perturbations, we induce diversity to disentangle the generation process from any potential training bias \citep{alzahrani2024benchmarks,carlini2021extracting}. The \textbf{Query Perturbator} is a \texttt{gpt-3.5-turbo} LLM agent 
that generates $n=5$ perturbations to the original query $q_0$ while retaining the same semantic meaning. 
In effect, the generation process can be summarized as returning a set of $\mathcal{Q}=\left\{q_0, q_1, \ldots, q_n\right\}$ query perturbations of size $n+1$. 
The \textbf{Query Perturbator's} singular purpose is to: \texttt{Rewrite the query in \{$n$\} radically different ways}. One prompt call is sufficient to discourage duplicates. Temperature is set to $1.0$ to prioritize creativity and lexical diversity.
Our analysis in Table \ref{tab:aggregate_experiments} shows that introducing perturbations, rather than sampling $n+1$ outputs for query $q_0$, results in a $13$ point spread between the lower and upper bound accuracy, a $12.5$ point decrease in Fleiss's $\kappa$ for agreement, while the modal accuracy remains largely unchanged. This suggests that perturbations inject variability into our Monte Carlo simulation, which is critical for observing diverse outputs and hallucinations. This corroborates the observation by \citet{alzahrani2024benchmarks} that LLMs are highly sensitive to even minor details.

\textbf{What is an Output Generator?}
\label{sec:outputgenerator}
For the perturbed set $\mathcal{Q}$ for a sample $q_0$, the \textbf{Output Generator} consists of $|\mathcal{Q}| = n + 1$ \textbf{six independent} \texttt{gpt-3.5-turbo} LLM agents to generate outputs $a_i \in \mathcal{A}$ for each variation $q_i \in \mathcal{Q}$. The LLM agent will receive (1) for {\color{BrickRed}\textbf{Extractive}} queries, a prompt with the query $q_i$, alongside context $c_i$, (2) for {\color{ForestGreen}\textbf{Multiple-Choice}} queries, candidate choices $k_i \in \mathcal{K}$, and (3) for {\color{Mulberry}\textbf{Abstractive}} queries, no additional context. Table \ref{tab:prompts} outline's each experiment's prompt procedure. Temperature for all experiments is set to $1.0$ to stress-test and encourage diversity.

\textbf{How Do We Measure Accuracy?}
\label{sec:acc}
\textit{Accuracy} serves as the measure of correctness, comparing the generated output $a_i$ to the ground truth $y$, using partial, case-insensitive matching with the \texttt{TheFuzz} library.
For {\color{ForestGreen}\textbf{Multiple Choice}} queries, the choice label is also considered. If there is no match between the output $a_i$ and the ground truth $y$, we assign $\mathbb{I}\bigl[a_i\not=y\bigr] \mapsto 1$; otherwise, $\mathbb{I}\bigl[a_i=y\bigr] \mapsto 0$. The results are compared to the baseline (original query $q_0$, output $a_0$), the mode (most common $a_i$), the lower bound (all correct), and the upper bound (at least one $a_i$ correct).

\textbf{How Do We Measure Agreement?}
\textit{Accuracy} alone is insufficient for evaluating the \textit{agreement} among multiple agents. 
To assess agreement, we report statistical measures for our Monte Carlo experiments including Item Difficulty ($\mathbf{\mu_D}$) \citep{Lord1952}, Fleiss's Generalized $\mathbf{\kappa}$ \citep{doi:10.1177/001316446002000104, Fleiss1971}, Mean Certainty / Entropy ($\mathbf{H_{\eta}}$) \citep{6773024, 4d25ef96-6507-346e-8e18-720c9de36b78}, and Gibbs' $\mathbf{M_2}$ Index \citep{10.1093/sf/53.3.468}.
These metrics help evaluate agreement levels amongst independent agents. For instance, high agreement on an incorrect answer indicates a misconception, while low agreement could suggest confusion or a poorly formulated query. To address this limitation in HalluciBot, we introduce a dual cross-entropy loss based on hallucination rates and consensus to improve the model's ability to distinguish good queries from bad queries.

\begin{table*}[!th]
\centering
\small
\begin{tabular}{llr|cc|cc|cccc} 
\hline
\multicolumn{3}{c|}{\textbf{Scenario}} & \multicolumn{4}{c|}{\textbf{Accuracy}} & \multicolumn{4}{c}{\textbf{Agreement}} \\ 
\hline
 & \textbf{Experiment} & \textbf{\#} & \textbf{Base} $\mathbf{\uparrow}$ & \textbf{Mode} $\mathbf{\uparrow}$ & \textbf{Lower} $\mathbf{\uparrow}$ & \textbf{Upper} $\mathbf{\uparrow}$ & $\mathbf{\mu_D}$ $\mathbf{\uparrow}$ & $\mathbf{H_{\eta}}$ $\mathbf{\uparrow}$ & $\mathbf{M_2}$ $\mathbf{\uparrow}$ & $\mathbf{\kappa}$ $\mathbf{\uparrow}$ 
 \\
\hline 
\multirow{4}{*}{\rotatebox[origin=c]{90}{\textbf{SINGLE}}} 

& \color{BrickRed} \textbf{Extractive} & 85,734 & 89.8 &	\textbf{90.3} &	83.6 &	94.6 &	89.8 &	91.4 &	92.0 &	90.4  \\

& \color{ForestGreen} \textbf{Multiple Choice} & 80,813 & 74.0 &	\textbf{75.8} &	58.1 &	88.0 &	73.8 &	90.3 &	83.7 &	75.5  \\

& \color{Mulberry} \textbf{Abstractive} & 200,693 & 56.2	& \textbf{56.7}	& 44.2	& 67.4	& 56.1	& 93.2	& 89.8	& 80.2 \\
\cline{2-11} 

& \textbf{Total} & 367,240 & 68.0 & \textbf{68.7} & 56.4 & 78.3 & 67.9 & 94.4 & 90.2 & 81.5 \\

\hline

\multirow{4}{*}{\rotatebox[origin=c]{90}{\textbf{MULTI}}} & \color{BrickRed} \textbf{Extractive} & 85,892 & \textbf{92.1} & 91.0 & 69.0 & 97.4 & 87.2 & 85.5 & 84.3 & 75.3 \\ 
& \color{ForestGreen} \textbf{Multiple Choice} & 81,697 & 76.3 & \textbf{76.8} & 47.4 & 91.6 & 71.8 & 75.2 & 71.3 & 61.9   \\ 
& \color{Mulberry} \textbf{Abstractive} & 202,248 & \textbf{55.9} & 53.9 & 32.9 & 67.3 & 51.2 & 81.5 & 80.0 & 69.1  \\ 
\cline{2-11} 

& \textbf{Total} & 369,837 & \textbf{68.6} & 67.4& 44.3 & 79.4 & 63.9 & 81.0 & 79.1 & 69.0   \\ 
\hline





\end{tabular}
\vspace{-1.25mm}
\caption{\label{tab:aggregate_experiments} 
Comparing \textbf{Single Query, Multiple Outputs (SINGLE)} vs. \textbf{Single Query, Multiple Perturbations, Single Output (MULTI)} Monte Carlo Experiments (\S\ref{sec:single_v_multi}). The reported metrics (\S\ref{sec:acc}) are calculated across all examples, regardless of the original dataset split. 
For the majority of scenarios, the SINGLE strategy outperforms the the MULTI approach in eliciting correct answers. 
Therefore, the SINGLE approach demonstrates higher agreement and tighter accuracy bounds, while the MULTI approach introduces more diverse responses and hallucinations with negligible impact on modal accuracy, allowing our simulation to generate more useful labels regarding query quality compared with a SINGLE approach. 
}
\vspace{-2.25mm}
\end{table*}

\begin{table*}[!th]
\centering
\small
\begin{tabular}{l|ccc|ccc|ccc|ccc}
\hline
\multicolumn{1}{c}{} & \multicolumn{3}{c|}{\textbf{Accuracy} $\mathbf{\uparrow}$} & \multicolumn{3}{c|}{\textbf{F1 Score} $\mathbf{\uparrow}$} & \multicolumn{3}{c|}{\textbf{Precision} $\mathbf{\uparrow}$} & \multicolumn{3}{c}{\textbf{Recall} $\mathbf{\uparrow}$} \\
\cline{2-13}
\multicolumn{1}{l}{\textbf{Model}} & \textbf{Train}  & \textbf{Val}  & \textbf{Test} 
& \textbf{Train}  & \textbf{Val}  & \textbf{Test} 
& \textbf{Train}  & \textbf{Val}  & \textbf{Test}  
& \textbf{Train}  & \textbf{Val}  & \textbf{Test}  \\
\hline 

\texttt{RoBERTa-base}        & 74.7 & 64.1 & 66.1  &  73.3 & 66.5 & 69.6 & 85.1 & 78.0 & 74.4 & 64.4 & 57.9 & 65.3 \\

\texttt{+ Scenario}          & 79.8 & 73.0 & 69.0  &  79.3 & 76.8 & 71.7 & \underline{\textbf{88.8}} & \textbf{\underline{81.5}} & \textbf{\underline{78.4}} & 71.5 & 72.6 & 66.0  \\

\texttt{+ Consensus} & 79.3 & 73.0 & 68.7 & 79.1 & 77.0 & 71.5 & 87.2 & 81.0 & 77.7 & 71.4 & 73.3 & 66.2 \\

\texttt{+ Calibration} & 80.3 & \underline{\textbf{73.6}} & \textbf{\underline{69.5}} & 81.4 & 78.8 & 73.6 & 83.6 & 78.4 & 75.6 & 79.2 & 79.2 & 71.7 \\

\texttt{\quad+ $\tau=0.341$} & \underline{80.4} & \underline{\textbf{73.6}} & \underline{\textbf{69.5}} & \underline{81.6} & \textbf{\underline{80.2}} & \textbf{\underline{76.0}} & 74.7 & 72.9 & 70.3 & \underline{90.0} & \underline{89.0} & \textbf{\underline{82.6}} \\

\hline

\texttt{RoBERTa-large} & & & & & & & & & & & \\



\texttt{+ Calibration} & 84.7 & 73.5 & 69.2 & \textbf{\underline{85.5}} & 78.5 & 73.0 & \underline{88.1} & \underline{78.9} & \underline{76.1} & 83.1 & 78.2 & 70.1 \\

\texttt{\quad+ $\tau=0.326$} & \underline{\textbf{84.8}} & \underline{\textbf{73.6}} & \underline{69.4} &83.5 & \underline{80.0} & \underline{75.6} & 75.0 & 71.8 & 70.5 & \underline{\textbf{94.2}} & \underline{\textbf{90.4}} & \underline{81.6} \\
\hline 
\end{tabular}
\vspace{-1.25mm}
\caption{\label{tab:HalluciBot-eval-binary}
HalluciBot Binary Evaluation Statistics. We report the Accuracy, F1, Precision, and Recall for all data splits. Probability threshold $\tau$ is computed along the closed interval $[0, 1]$ in increments of $0.001$ to maximize the validation F1 score for the final model. The best ablation per base model is \underline{underlined}, while the overall best performing model is in \textbf{bold}. 
}
\vspace{-2.25mm}
\end{table*}

\subsection{Converting Monte Carlo Estimates To Labels}
\label{sec:make_labels}

\textbf{Empirical Estimate.} The \textit{probability of hallucination} for a query $q_0$, denoted as $p_h(q_0)$, can be empirically estimated based on the output $a_i \in \mathcal{A}$ of our Multi-Agent Monte Carlo simulation. We define the indicator function $\mathbb{I}$ to measure the incorrectness of an output $a_i$ with respect to the ground truth $y$ for query $q_0$.
\begin{equation*}
    p_h\bigl(q_0\bigr) \approx \frac{1}{n+1}\sum \mathbb{I}\big[a_i \not= y\big]
\end{equation*}

\textbf{Binary Hallucination \& Consensus Labels.}
To assess the propensity to hallucinate, we simplify the problem by considering two response values: \textit{whether $\mathit{q_0}$ produces any hallucination or not}. Thus, we define the binary values for the probability of any hallucination as $p_b(q_0)$. Furthermore, we craft a secondary \textbf{consensus} label $p_c(q_0)$ that is a proxy for the agreement of the query. It maps the set of unique answers to $1$ if there is any disagreement, otherwise we assign $0$ for perfect agreement ($1$ unique answer). Therefore, we can train a 2 head output \texttt{Consensus} model to predict the hallucination probability $p_b(q_0)$, and if the query will cause confusion or consensus $p_c(q_0)$.
\begin{align*}
    p_b\bigl(q_0\bigr) &= 
        \begin{cases}
            1 & \text{if } p_h\bigl(q_0\bigr) > 0 \\
            0 & \text{if } p_h\bigl(q_0\bigr) = 0 \\
        \end{cases} \\
    p_c\bigl(q_0\bigr) &= 
    \begin{cases}
        1 & \text{if } \bigl| \{a_i \,|\, a_i \in \mathcal{A} \} \bigr| > 1 \\
        0 & \text{if } \bigl| \{a_i \,|\, a_i \in \mathcal{A} \} \bigr| = 1 \\
    \end{cases}
\end{align*}

\subsection{How To Train a Classifier?}
\label{sec:train}
Once the Monte Carlo simulation is complete for our training corpus composed of
369,837 queries spanning 13 different datasets  (Appendix \ref{app:datasets}, Tables \ref{tab:quick-ds} \& \ref{tab:overview}), we start training our classifier. These queries encompass {\color{BrickRed} \textbf{Extractive}}, {\color{ForestGreen} \textbf{Multiple Choice}}, and {\color{Mulberry} \textbf{Abstractive}} scenarios.  Each scenario, with or without additional context, affects the hallucination rate of \texttt{gpt-3.5-turbo}.
These simulated estimates are directly proportional to the approximated rates of hallucination $p_h$. 
\begin{itemize}[noitemsep, leftmargin=*, topsep=0pt, partopsep=0pt, label={\tiny\raisebox{0.5ex}{$\blacktriangleright$}}]
    \item With a synthetic labeled set of queries $q_0$ and their rate of hallucinations $p_{h}(q_0)$, we train an encoder-style \texttt{RoBERTa} \citep{DBLP:journals/corr/abs-1907-11692} classifier to estimate the hallucination probability density from our Monte Carlo simulation.
    \item We ablate two versions: a binary model to estimate the propensity a query can hallucinate, and a consensus-aware model to also predict the expected agreement of outputs if sampled $n+1$ times. 
\end{itemize}
Our experiments constrain the number of perturbations to $n=5$, and when including the original query and output, we can model the hallucination rate for $n+1=6$ modes. This translates to increments of $16.\overline{6}\%$ in hallucination rates. 

\begin{figure*}[!t]
\centering
\includegraphics[clip, trim={0.7cm 0.5cm 0.65cm 0.71cm}, height=3.4cm]{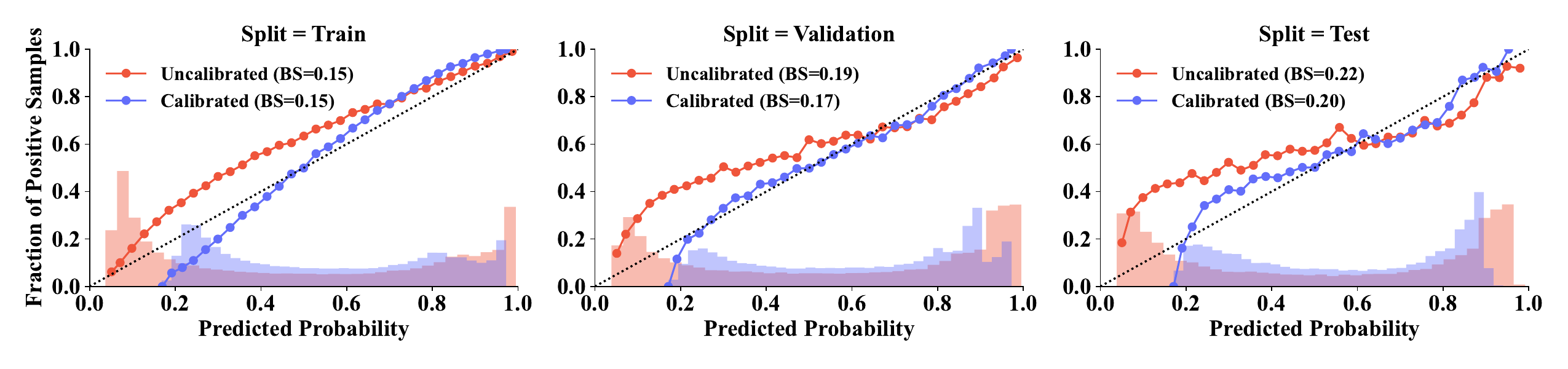}
\includegraphics[clip, trim={1.09cm 2.45cm 2.45cm 1.92cm}, height=2cm]{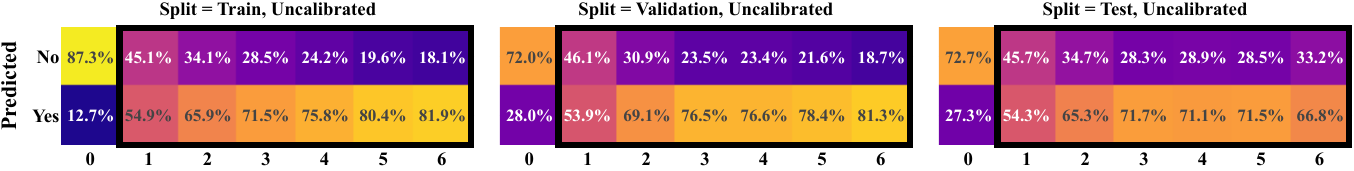}
\includegraphics[clip, trim={1.09cm 2.15cm 2.45cm 1.92cm}, height=2.205cm]{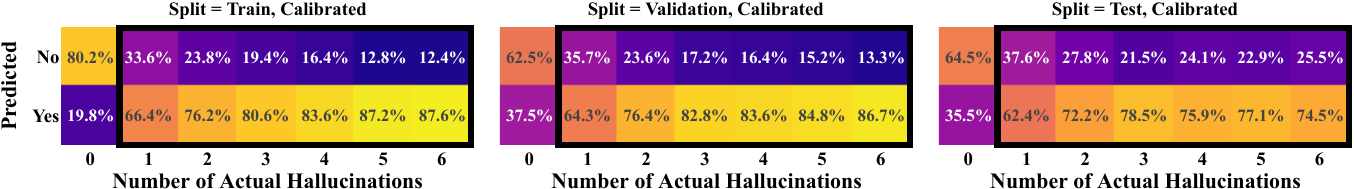}
\caption{Top: HalluciBot calibration curves with Brier Scores (BS), alongside the histogram of predicted probabilities. Bottom: Predicted hallucination labels juxtaposed against observed hallucination rates during our Monte Carlo simulation, with calibrated matrix below. We highlight $1$-$6$ as corresponding to the binary label ``Yes - Hallucinatory'' ($y=1$) during training. Notably, there is significant confusion in queries that are borderline ($1$, $2$) rather than majority hallucinatory prone ($3$-$6$).
}
\label{fig:figure-confusion}
\vspace{-2mm}
\end{figure*}

\textbf{How To Encode a Query's Scenario?}
We conduct an ablation study to explore if incorporating the query's scenario mitigates hallucinations. To create the prompt, we prepend the original query $q_0$ with either \texttt{\color{BrickRed}\textbf{[EXTRACTIVE]}}, \texttt{\color{ForestGreen}\textbf{[MULTIPLE CHOICE]}}, or \texttt{\color{Mulberry}\textbf{[ABSTRACTIVE]}}, using the format \texttt{<<\{tag\} \{$\mathtt{q_0}$\}>>}. Our hypothesis is based on recent research that highlights the use of RAG \cite{yan2024corrective,NEURIPS2020_6b493230,pmlr-v119-guu20a} to alleviate hallucinations. The additional context provides valuable signals related to the hallucination rate of the original query. Furthermore, we apply this technique to distinguish our experimental results from reused datasets in different scenarios, such as SciQ \citep{SciQ} and SQuADv2 \citep{rajpurkar2016squad, rajpurkar2018know}. 


\textbf{H4R: Downstream Modes For HalluciBot.}
Without HalluciBot's feedback, typical query rewriting models have to act as both an implicit critic and a generator. As a proxy reward model, HalluciBot's probabilistic feedback on a query's quality, given the dual prediction heads for hallucination and consensus, can guide a query rewriting process using an independent \texttt{gpt-3.5-turbo} LLM, \textit{before output generation}. In essence, HalluciBot provides the following downstream modes for handling potentially hallucinatory queries:
\begin{enumerate}[noitemsep, leftmargin=*, topsep=0pt, partopsep=0pt, ]
    \item \textbf{Rewrite Mode:} A single-shot iterative rewrite of queries classified as hallucinatory.
    \item \textbf{Rank Mode:} Generating $N$ intermediate perturbations, sorted by HalluciBot's class probabilities for fine-grained scoring. In our implementation, the number of outputs were controlled by the number of chat completion choices in \texttt{gpt-3.5-turbo}'s API call.
    \item \textbf{Route Mode:} For \textbf{\color{Mulberry}Abstractive} or \textbf{\color{BrickRed}Extractive} queries classified as hallucinatory, testing if switching the scenario (e.g. between RAG or direct inference) generates more robust classifications and generations.
\end{enumerate}
The query rewriting prompt is found in Appendix Listing~\ref{lst:query_rewrite_prompt}.

\begin{figure*}[t]
\begin{minipage}[t]{.69\textwidth}
\centering
\vspace{0pt}
\small
\begin{tabular}{l|r|l|r|l|r} 
\hline
\multicolumn{1}{c}{\textbf{Metrics ($\%$)}}   
& \textbf{Test} 
& \multicolumn{1}{c}{\textbf{Metrics ($\%$)}}   
& \textbf{Test} 
& \multicolumn{1}{c}{\textbf{Metrics ($\%$)}}   
& \textbf{Test}\\
\hline
\multicolumn{2}{c|}{\cellcolor{LightGrayCell}}
& \multicolumn{2}{c|}{\cellcolor{MiddleCell} (B) [HB] }
& \multicolumn{2}{c}{\cellcolor{WhiteSmokeCell} (C) [HB]}
\\
\multicolumn{2}{c|}{\multirow{-2}{*}{\cellcolor{LightGrayCell} (A) Naive Rewrite}}
& \multicolumn{2}{c|}{\cellcolor{MiddleCell}Informed Single Rewrite}
& \multicolumn{2}{c}{\cellcolor{WhiteSmokeCell}Best-of-N Rewrite}
\\\hline
$+$ Class Transitions  &	6.5 
& $+$ Class Transitions   &	30.2
& $+$ Class Transitions   &	50.6
\\
$-$ Class Transitions &	3.2 
& Rewrite Accuracy	& 	
& Rewrite Accuracy	& 	
\\
Unneeded Rewrites & 46.6 
& \enspace~Top-5  & \textbf{\color{ForestGreen}94.3}
& \enspace~Top-5  & \textbf{\color{ForestGreen}95.2}
\\
$$	& $$
& \enspace~Similarity Score  & \textbf{\color{Mulberry}46.9}
& \enspace~Similarity Score  & \textbf{\color{Mulberry}47.4}
\\
\hline
\multicolumn{2}{c|}{\cellcolor{LightGrayCell} (D) Assuming HB Ratiocinate}
& \multicolumn{2}{c|}{\cellcolor{MiddleCell} (E) [HB w/ Consensus]}
& \multicolumn{2}{c}{\cellcolor{WhiteSmokeCell} (F) [HB w/ Consensus]}
\\
\multicolumn{2}{c|}{\cellcolor{LightGrayCell}Naive Rewrite (Baseline)}
& \multicolumn{2}{c|}{\cellcolor{MiddleCell}Informed Single Rewrite}
& \multicolumn{2}{c}{\cellcolor{WhiteSmokeCell}Best-of-N Rewrite}
\\\hline
$+$ Class Transitions  &	14.8 
& $+$ Class Transitions   &	31.9
& $+$ Class Transitions   &	51.4
\\
Rewrite Accuracy	& 	
& Rewrite Accuracy	& 	
& Rewrite Accuracy	& 	
\\
\enspace~Top-5 & \textbf{\color{ForestGreen}92.9}
& \enspace~Top-5 & \textbf{\color{ForestGreen}90.2}
& \enspace~Top-5 & \textbf{\color{ForestGreen}95.7}
\\
\enspace~Similarity Score  & \textbf{\color{Mulberry} 41.7}
& \enspace~Similarity Score  & \textbf{\color{Mulberry} 57.5}
& \enspace~Similarity Score  & \textbf{\color{Mulberry} 55.9}
\\
\hline
\end{tabular}
\vspace{-1.75mm}
\captionof{table}{Query generation metrics under each HalluciBot (HB) strategy. 
 \textbf{\color{ForestGreen}Multiple Choice} queries were evaluated on a soft accuracy criterion where the score is +1 if any of the $n$ generations match the ground truth. For \textbf{\color{Mulberry}Abstractive} queries the average cosine similarity score between the ground truth and the $n$ generation outputs is reported. The embedding vectors for similarity computation are obtained using \texttt{all-MiniLM-L6-v2} \citep{wang2020minilm, reimers-2019-sentence-bert}.}
\label{tab:query_gen}
\end{minipage}\hfill
\begin{minipage}[t]{.29\textwidth} 
    \centering
    \vspace{0pt}
    \includegraphics[trim={0.27cm 0.305cm 0.23cm 0.27cm}, clip, width=\textwidth]{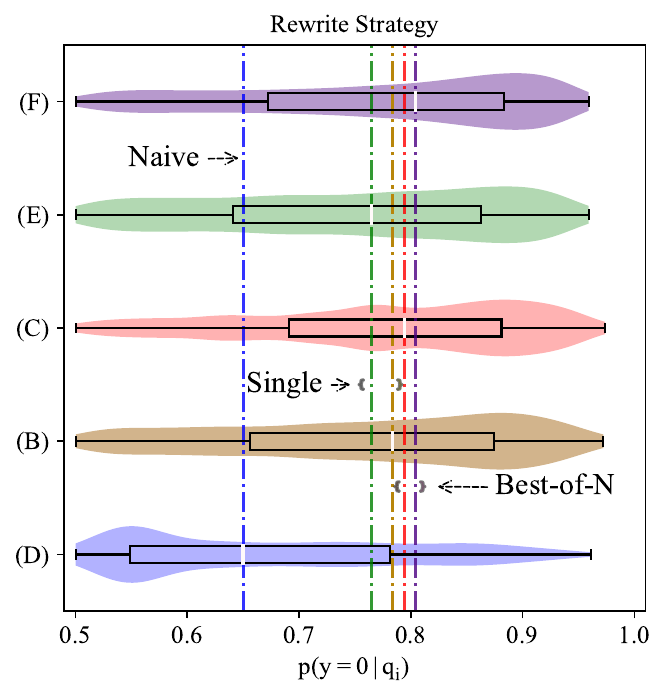}
    \vspace{-1.75mm}
    \captionof{figure}{Class probability of queries that were rewritten and reclassified to be non-hallucinatory.}\label{fig:figure-violin}
\end{minipage}
\vspace{-2mm}
\end{figure*}

\vspace{-1mm}
\section{Experimental Setup}
\label{experiment}

\textbf{Dataset Coverage \& Scenario Split.}
\label{sec:dataset}
Our experiments include 13 datasets (Table \ref{tab:quick-ds}) divided into 3 scenarios: {\color{BrickRed}\textbf{Extractive}}, {\color{ForestGreen}\textbf{Multiple Choice}}, and {\color{Mulberry}\textbf{Abstractive}}. To evaluate the impact of context, we use SQuADv2 \citep{rajpurkar2016squad,rajpurkar2018know} to simulate RAG \citep{NEURIPS2020_6b493230, pmlr-v119-guu20a}. To assess the effect of multiple choice queries, we repurposed TruthfulQA \citep{lin-etal-2022-truthfulqa} and SciQ \citep{SciQ} for two experiments: one where the output agents select from the choices or context, and another where LLM agents generate outputs without context. We maintain the original train, validation, and test splits across scenarios to prevent information leakage to HalluciBot.
Prompt templates for each \texttt{gpt-3.5-turbo} agent can be found in App. Table \ref{tab:prompts}. All LLM agents share the same set of parameters, as described in App. Table \ref{tab:gpt_params}.

\textbf{HalluciBot Training Parameters \& Environment.}
We employed HuggingFace's Trainer class with the Adam optimizer \citep{kingma2017adam} for training, reporting efficiency and training times in App. Table \ref{tab:HalluciBot-train}. All experiments were conducted on an AWS EC2 instance with a single GPU (App. Table \ref{tab:aws_interface}). 
HalluciBot is fine-tuned from both pretrained \texttt{BERT} \citep{DBLP:journals/corr/abs-1810-04805} and \texttt{RoBERTa} \citep{DBLP:journals/corr/abs-1907-11692} models (App. Table \ref{tab:HalluciBot_params}). To address label imbalance, we employed a weighted class loss where each class weight is assigned to its inverted frequency in the training set. 
The train, validation, and test splits follow the original divisions of the datasets. Specifically, there are 302,492 training, 44,491 validation, and 22,854 testing samples. The distribution of labels across these splits is summarized in Table \ref{tab:splits}, and fine-grained splits per set are enumerated in App. Table \ref{hallucination_rates}. We also apply Platt calibration \citep{vasilev2023calibration, guo2017calibration, 10.1145/1102351.1102430, Platt1999ProbabilisticOF} based on the validation logits to help ensure that the raw probabilities align better with the true class labels (Figure~\ref{fig:figure-confusion}).

\section{Analysis \& Discussion}
\label{sec:analysis}

\textbf{Ablation: Perturbations Induce Output Diversity.}
\label{sec:single_v_multi}
We examine the impact of perturbations on the robustness of \texttt{gpt-3.5-turbo} in question-answering tasks by comparing two strategies: \textbf{Single Query, Multiple Outputs (SINGLE)} and \textbf{Single Query, Multiple Perturbations, Single Output (MULTI)}. In the SINGLE strategy, we sample $n+1$ outputs from the original query $q_0$. In the MULTI strategy, $n$ perturbations of the original query $q_0$ are used, and each perturbation $q_i$ is answered once.
Table \ref{tab:aggregate_experiments} shows that while baseline accuracy remains consistent, the lower bound accuracy drops by $12.1$ points in the MULTI setting. Additionally, agreement metrics, as indicated by Fleiss's $\kappa$, decrease by $12.5$ points, indicating reduced consistency.
In summary, (1) the SINGLE strategy results in higher agreement and lower-bound accuracy while (2) the MULTI strategy increases response diversity and hallucination rates but offers a slight improvement in upper-bound accuracy for \textbf{\color{BrickRed}{Extractive}} and \textbf{\color{ForestGreen}{Multiple Choice}} scenarios. This suggests that perturbations can enhance query quality by introducing necessary diversity, despite minor variations in modal accuracy.

\textbf{Ratiocinate: Can HalluciBot Detect Hallucinatory Queries?}
Differentiating the scenario in HalluciBot's prompt yielded a strong $+10.3\%$ increase in validation F1 score. The calibrated, threshold-tuned \texttt{RoBERTa-base} HalluciBot in Table \ref{tab:HalluciBot-eval-binary} achieves a test accuracy of $69.5\%$ with a macro F1-score of $76.0\%$. Further breaking down the results in Figure \ref{fig:figure-confusion}, calibrating our models with Platt scaling improves the discriminating power for borderline queries, where the observed number of hallucinations was minimal ($y\in\{1, 2\}$). Finally, HalluciBot demonstrates strong recall scores ($89.0\%$ validation, $82.6\%$ testing) to effectively flag risky queries that are likely to generate \textit{at least one hallucination} during inference. The importance of HalluciBot as a ratiocinating process can be seen in [Table~\ref{tab:query_gen}~(A)] under a naive rewriting strategy. Without HalluciBot, a naive rewrite strategy has the potential to convert queries originally estimated to be non-hallucinatory to hallucinatory (negative class transition), because there is no mechanism to differentiate queries. With HalluciBot restricting the test set to only potentially hallucinatory queries (11.2K samples), a naive rewrite [Table~\ref{tab:query_gen} (D)] can only enact positive class transitions ($+14.8\%$), converting queries originally estimated to hallucinate to non-hallucinatory. Furthermore, HalluciBot acting as an arbitrator can prevent computationally expensive rewrite calls for $46.6\%$ of the test set (10.2K samples deemed to be non-hallucinatory).


\textbf{Rewrite: As a Feedback Mechanism.}
HalluciBot's feedback allows us to generate a more informed query [Table~\ref{tab:query_gen}~(B)] resulting in better class transition probabilities than an uninformed rewrite strategy [Table~\ref{tab:query_gen}~(D)]. This translates to a $14.8\%$ positive class transition and a $1.4\%$ increase in \textbf{\color{ForestGreen}Multiple Choice} accuracy as well as $5.2\%$ improvement in generation similarity for \textbf{\color{Mulberry}Abstractive} queries. Utilizing consensus information during the query rewriting process [Table~\ref{tab:query_gen}~(E)] generates a slightly larger positive class transition ($31.9\%$ vs. $30.2\%$) than without [Table~\ref{tab:query_gen}~(B)]. 

\textbf{Rank: Best-of-N Rewrite.} 
A Best-of-N  rewrite strategy [Table~\ref{tab:query_gen}~(C, F)] demonstrates a $19.5\%$ and $20.4\%$ gain in positive class transitions in both experiments over a single rewrite [Table~\ref{tab:query_gen}~(B, E)]. Therefore, HalluciBot's estimated probabilities can be used as a proxy reward model when ranking $n$ sample perturbations. The violin plots in Figure~\ref{fig:figure-violin} shows that HalluciBot is able to select better queries in the Best-of-N rewrite setting than with a single rewrite, with a higher median predicted non-hallucinatory probability ($79.5\%$ (C) vs. $78.4\%$ (B); $80.4\%$ (F) vs. $76.5 \%$ (E)). 
This means that HalluciBot evaluated the rewritten queries to be non-hallucinatory with greater probability. 
All rewrites are single-shot without subsequent tuning or iterations.

\textbf{Route: {\color{Mulberry}Abstractive} to {\color{BrickRed}Extractive}.}
The test set had 948 {\textbf{\color{Mulberry}Abstractive}} queries that were classified to be hallucinatory. Conditioned on this information, switching the scenario to {\textbf{\color{BrickRed}Extractive}} results in a $+60.0\%$ positive class transition. In contrast, a rewriting process without a scenario change has a much smaller  $9.7\%$ class transition. 
With HalluciBot's ability to distinguish scenarios, it can help determine whether direct inference or RAG is more effective for a particular query. 

\section{Conclusion}
\label{sec:conclusion}
We propose a heretofore relatively unexplored realm of hallucination mitigation - predicting a query's hallucination probability. 
HalluciBot empirically estimates how the query itself may induce hallucination and its training corpus consists of diverse scenarios and domains to ensure robustness. 
Institutions can implement HalluciBot to measure user accountability and improve the robustness of LLM's performance via our H4Rs (\textbf{``Ratiocinate''}, \textbf{``Rewrite''}, \textbf{``Rank''}, \textbf{``Route''}). Thus, HalluciBot's academic and practical contributions add to the ever-growing concerted effort of enabling a robust language generation ecosystem for society.  

\textbf{Limitations.}
HalluciBot relies on automated LLM crowdsourcing via Monte Carlo sampling to generate perturbations, which may introduce noise. Sampling is computationally expensive during training, but is balanced by its ability to curate large, diverse corpora spanning various domains and scenarios. HalluciBot can be trained on a mixture of LLMs or used as a proxy-reward model for RL-tuned generators. 

\section*{Disclaimer}
{
\small
This paper was prepared for informational purposes by the Artificial Intelligence Research group of JPMorgan Chase \& Co. and its affiliates ("JPMorgan'') and is not a product of the Research Department of JPMorgan. JPMorgan makes no representation and warranty whatsoever and disclaims all liability, for the completeness, accuracy or reliability of the information contained herein. This document is not intended as investment research or investment advice, or a recommendation, offer or solicitation for the purchase or sale of any security, financial instrument, financial product or service, or to be used in any way for evaluating the merits of participating in any transaction, and shall not constitute a solicitation under any jurisdiction or to any person, if such solicitation under such jurisdiction or to such person would be unlawful.
}

\bibliography{aaai25}

\clearpage


\appendix

\section{Definitions}

\subsection{What is Extractive QA?}
Extractive Question-Answering involves extracting answers directly from a given context. It can be accomplished using span-detection encoders that predict the start and end tokens of the relevant portion in the context \citep{DBLP:journals/corr/abs-1810-04805}. Another approach is Retrieval-Augmented Generation (RAG), where the model generates the answer based on a selected passage containing the necessary information \citep{NEURIPS2020_6b493230,pmlr-v119-guu20a}.

\subsection{What is Multiple Choice QA?}
Multiple Choice Question-Answering is a task in few-shot learning where a model is given a question and a set of answer choices. Usually, one of the choices is correct, while the rest are distractors. In an encoder-based approach, the model replicates the question for each choice, generates a scalar response, and applies a Softmax across the choices to determine the best answer \citep{DBLP:journals/corr/abs-1810-04805}. In a generative setting, the model can generate an answer choice instead of selecting from the given options, using few-shot prompting techniques \citep{wei2022finetuned}.

\begin{figure}[t]
\centering
\includegraphics[trim={0.4cm 0.23cm 0.45cm 0.12cm},clip,width=\columnwidth]{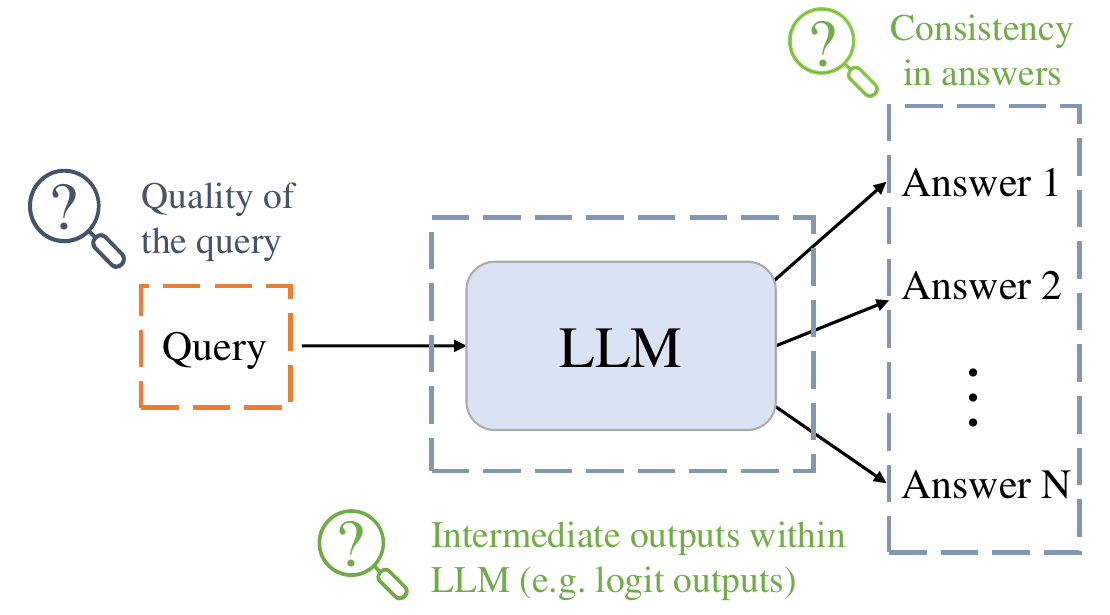}
\caption{
To assess the possibility of hallucination, existing literature focuses on either the outputs or the intermediate outputs within the model. Our method instead focuses on directly assessing the quality of the query, which is quantified as as how likely it will lead to a hallucination.} 
\label{fig:teaser}
\end{figure}

\subsection{What is Abstractive QA?}
Abstractive Question-Answering refers to the generation of answers without access to context or candidates. Also known as Closed-book Generative QA, where no context is provided, abstractive methods are used to generate answers based solely on the question \citep{wei2022finetuned}. Encoder models can be employed in this approach, which convert commonly occurring answer tokens into class labels \citep{kim2021vilt}. A multi-class model is then trained using Softmax regression to predict the answer from the question.

\subsection{What is Temperature and Nucleus Sampling?}
Decoder-based language models approximate the distribution of the next probable token across a given vocabulary $V$. This is often implemented by applying the Softmax function across a language model's final output vector $u_i$, per token $x_i$. 

$$p\bigl(y_i\big|x_{1:i-1}\bigr) = \frac{e^{u_i}}{\sum_j e^{u_j}}$$
Greedy decoding takes the most likely predicted token of the output distribution per sequence step, but often yields poor results and a lack of diversity. To combat this, alternative approaches leverage multinomial sampling across the distribution to sample less likely tokens. Temperature $T$ is used to control the smoothing of the distribution. As $T\to\infty$, the output distribution is smoother and more uniform; therefore unlikely tokens become more likely to sample. As $T\to0$, the distribution approaches a Kronecker Delta ($\delta_{ij}$) function, centered on the token with the most probability mass (and mimics greedy decoding strategies). Temperature is often implemented as an adjusted Softmax function \citep{holtzman2020curious}.

$$p\bigl(y_i\big|x_{1:i-1}\bigr) = \frac{e^{u_i/T}}{\sum_j e^{u_j/T}}$$
However, sampling across the full vocabulary can lead an LLM to produce extremely unlikely generations. To restrict the space of possible outcomes, two approaches exist: Top-$K$ Sampling and Nucleus Sampling. Top-$K$ sampling is implemented to include tokens that are in the Top-$K$ most probable tokens \citep{fan-etal-2018-hierarchical,holtzman-etal-2018-learning,radford-topk}. One limitation is the static window of possible tokens considered, where unlikely tokens can still be considered even if most of the probability mass is distributed amongst fewer tokens than $K$.

Nucleus Sampling, instead of considering the Top-$K$ most probable tokens in vocabulary $V$, only considers the smallest set of tokens that have a cumulative likelihood greater than some threshold $p_i$. Therefore, each decoding step will consider the most likely Top-$p_i$ tokens $V^{(p_i)}\subset V$, automatically eliminating improbable tokens from being accidentally sampled through masking \citep{holtzman2020curious}.

$$\sum_{x\in V^{(p_i)}} p\bigl(x\big|x_{1:i-1}\bigr) \geq p_i$$

\subsection{What is a Multi-Agent Simulation?}
In a Multi-Agent simulation, independent agents collaborate and interact to find a solution \citep{10.5555/1671238}. In the context of our experiments, we have two types of agents: the \textbf{Query Perturbator} and $n+1$ \textbf{Output Generators}. The purpose of the Multi-Agent simulation is to generate independent outputs after perturbing the query $q_0$ to observe the rate of hallucination. The presence of $n+1$ queries and agents is to ensure a balanced representation of hallucination, preventing skewed results towards either extreme. The set of outputs generated by the agents is then evaluated against the ground truth value $y$ to estimate the hallucination rate. Through multiple samplings across all 369,837 queries, we train HalluciBot to accurately identify and assess the risk of hallucination.

\subsection{What is Monte Carlo Sampling?}
Monte Carlo sampling is a technique used to approximate unknown quantities of interest when it's difficult to derive an exact solution \citep{10.5555/2380985}. This is particularly useful when dealing with hidden latent variables that cannot be directly observed, such as the case of hallucination caused by interacting with a complex random variable (e.g., a Large Language Model).
To address this challenge, we use Monte Carlo sampling. The idea is to conduct a Multi-Agent simulation for each query in our training set. By doing so, we can estimate the probability of hallucination by observing the accuracy rate per query across lexical perturbations. The more simulations we conduct, the more accurate our estimation becomes.
In our experiments, we sample from a series of 369,837 Multi-Agent simulations that examine hallucination rates across perturbations for any query $q_0$. Therefore, we can estimate the hallucination rate $p_h(q_0)$ for query $q_0$, given the ground truth $y$ and set of answers $a_i\in \mathcal{A}$.

$$p_h\bigl(q_0\bigr) \approx \frac{1}{n+1}\sum \mathbb{I}\big[a_i \not= y\big]$$
By deriving labels to estimate $p_h(q_0)$, we train HalluciBot to approximate the true risk of hallucination.

\subsection{What is Ordinal Regression?}
Ordinal Regression, in contrast to Softmax regression, learns a sequence of cutpoints to divide the prediction space into classes, allowing models to bias errors to the nearest class label. This is useful when our labels have order, such as estimating the number of stars for a review or the expected rate of hallucination. Let $f(x)$ be an encoder-model that outputs a single scalar score, such that $f(x) \mapsto \mathbb{R}$. Additionally, let $\sigma(x)$ be the Sigmoid function:
\begin{equation*}\sigma(x) = \frac{1}{1+\exp{\left(-x\right)}}\end{equation*}
Therefore, the probability of a binary ordinal classifier centered at cutoff point $0$, to  differentiate in log-probability space the positive and negative outcomes, is:
\begin{align*}
    p\bigl(y=1\big|x\bigr) &= \sigma\bigl(f(x)\bigr) \\
    p\bigl(y=0\big|x\bigr) &= 1 - p\bigl(y=1\big|x\bigr) \\
    &= 1 - \sigma\bigl(f(x)\bigr) \\
\end{align*}
Expanding to several classes $K$, we can divide the probability space into $K-1$ cutpoints $c_1,\dots,c_{K-1}$ with the property that $c_1 < c_2 < \cdots c_{K-1}$, such that the probability for each class is defined as follows \citep{353a0d24-9c24-3a11-a330-afc86b9c39c8}:
\begin{align*}
    p\bigl(y=0\big|x\bigr) &= \sigma\bigl(c_1 - f(x)\bigr) \\
    & \,\,\, \vdots  \\
    p\bigl(y=k\big|x\bigr) &= \sigma\bigl( c_{k+1} - f(x)\bigr) \\
    & \quad\quad - \sigma\bigl(c_{k} - f(x)\bigr)  \\
    & \,\,\, \vdots  \\
    p\bigl(y=K-1\big|x\bigr) &= 1  - \sigma\bigl(c_{K-1} - f(x)\bigr) 
\end{align*}
To enforce the property $c_1 < c_2 < \cdots c_{K-1}$ while allowing our thresholds to be differentiable, we employ a cumulative sum on a set of unbounded, learnable parameters $\theta_1, \dots, \theta_{K-1}$, transformed by a Softplus ($\beta=1$) to avoid adding negative values; this ensures our thresholds are always increasing \citep{7280459}. Therefore, each cutpoint $c_k$ will be of the from:
\begin{align*}
\text{Softplus}(\theta_k) & = \frac{1}{\beta} \cdot \log \bigl(1 + \exp\left(\beta \cdot \theta_k\right)\bigr) > 0 \\\\
    c_1 & = \theta_1 \\
c_k & = \theta_1 + \sum_{i=2}^{k} \text{Softplus}(\theta_i) 
\end{align*}

\section{Additional Experiments \& Parameters}
\label{sec:experiment_details}

\subsection{LLM Agent Settings}
We outline the parameters for all our LLM agents in Table \ref{tab:gpt_params}. In addition, the prompts used per component and scenario are outlined in Table \ref{tab:prompts}.

\begin{figure*}[!th]
\begin{minipage}[t]{.47\linewidth}
\centering
\small
\begin{tabular}[t]{l|c}
\hline
\multicolumn{2}{c}{\textbf{LLM Parameters (All Agents)}} \\
\hline
Engine & \texttt{gpt-35-turbo-16k}\\
Version & \texttt{2023-09-01-preview}\\
Temperature & 1.0 \\
Frequency Penalty & 0.0 \\
Presence Penalty & 0.0 \\
Top P & 0.95 \\
Max Tokens & 800 \\
Stop & \texttt{None} \\
Seed & 123 \\
\hline
\end{tabular}
\captionof{table}{\label{tab:gpt_params}
Configuration for \texttt{gpt-3.5-turbo}. The same set of parameters were used for both the Query Perturbator and Output Generator.
}
\end{minipage}\hfill 
\begin{minipage}[t]{.47\linewidth}
\centering
\small
\begin{tabular}[t]{l|c}
\hline
\multicolumn{2}{c}{\textbf{Training Instance Parameters}} \\
\multicolumn{2}{c}{\textbf{(All HalluciBot Models)}} \\
\hline
Instance & \texttt{g4dn.4xlarge}\\
\# GPUs & \texttt{1}\\
GPU Type & \texttt{NVIDIA T4}\\
GPU Memory (GiB) & \texttt{16}\\
vCPUs & \texttt{16} \\
RAM (GiB) & \texttt{64} \\
\hline
\end{tabular}
\captionof{table}{\label{tab:aws_interface}
AWS EC2 training instance parameters for training HalluciBot.
}
\end{minipage}
\end{figure*}

\begin{table*}[!th]
\centering
\small
\begin{tabular}{l|l}
\hline
\multicolumn{2}{c}{\textbf{Prompt Templates}} \\
\hline
\multicolumn{1}{c|}{\textbf{Agent}} & \multicolumn{1}{c}{\textbf{Prompt}} \\
\hline 
{\color{WildStrawberry} \textbf{Query}}
& \texttt{Rewrite the query in \{$n$\} radically different ways.} \\
{\color{WildStrawberry} \textbf{Perturbator}} & \texttt{Query: \{$q_i$\}} \\
\hline
{\color{BrickRed} \textbf{Extractive}}
& \texttt{You will answer the user's query.} \\
{\color{BrickRed} \textbf{Output}} & \texttt{Context: \{$c_i$\}}\\
{\color{BrickRed} \textbf{Generator}} & \texttt{Query: \{$q_i$\}} \\
 & \texttt{Answer:} \\
\hline
\color{ForestGreen} \textbf{Multiple Choice} 
& \texttt{You will answer the user's query.} \\
\color{ForestGreen} \textbf{Output} & \texttt{Query: \{$q_i$\}}\\
\color{ForestGreen} \textbf{Generator}  & \texttt{A) \{$k_0$\}} \\
 & \texttt{\vdots} \\
 & \texttt{Z) \{$k_m$\}} \\
 & \texttt{Answer:} \\
\hline
 {\color{Mulberry} \textbf{Abstractive}}
  & \texttt{You will answer the user's query.} \\
 {\color{Mulberry} \textbf{Output}} & \texttt{Query: \{$q_i$\}} \\
 {\color{Mulberry} \textbf{Generator}} & \texttt{Short Answer:} \\
\hline

\end{tabular}
\caption{\label{tab:prompts}
Prompt templates for all \texttt{gpt-3.5-turbo} agents. Appendix \S\ref{app:datasets} outlines each dataset's taxonomy for output generation. For {\color{ForestGreen} \textbf{Multiple Choice}} experiments, we do not perturb any of the original choices $k_i \in \mathcal{K}$, and enumerate them in a consistent order for all perturbations $q_i \in \mathcal{Q}$. The output generator produces outputs $a_i \in \mathcal{A}$ for each perturbed query $q_i \in \mathcal{Q}$. For our experiments, the number of agents is set to $n=5$, yielding $n+1=6$ queries and outputs per example. For {\color{BrickRed}\textbf{Extractive}} output generation, context is denoted as $c_i$.
}\vspace{3mm}
\end{table*}

\subsection{Aggregated Monte Carlo Results}
The aggregated Monte Carlo results for the \textbf{MULTI} approach demonstrate the relative performance of different question-answering scenarios. 
On average, the results (Table \ref{tab:aggregate_experiments}) indicate that {\color{BrickRed} \textbf{Extractive}} outperforms {\color{ForestGreen} \textbf{Multiple Choice}}, which, in turn, outperforms {\color{Mulberry} \textbf{Abstractive}} when subjected to perturbations. This trend suggests that the performance of \texttt{gpt-3.5-turbo} is influenced by the presence of additional content.
{\color{Mulberry} \textbf{Abstractive}} tasks show the greatest variation in agent response under perturbations, highlighting the effectiveness of added context in mitigating hallucinations (Figures \ref{fig:figure-multi} \& \ref{fig:figure-binary}). Table \ref{results} provides a full breakdown.
\begin{itemize}[noitemsep, leftmargin=*, topsep=0pt, partopsep=0pt, label={\tiny\raisebox{0.5ex}{$\blacktriangleright$}}]
\item 
{\textbf{\color{BrickRed}Extractive:}} With context, \texttt{gpt-3.5-turbo} performs well on the SQuADv2 dataset. The mode accuracy ($91.0\%$) and agreement ($75.3\%$) of the agents are high. Even under radical perturbations, having \textit{unaltered} context provides robustness to the agent's capacity to answer correctly.
\item 
{\textbf{\color{ForestGreen}Multiple Choice:}}
Access to answer choices mitigates hallucinations across perturbations. The ensemble accuracy is slightly higher than the baseline ($+0.5\%$), showcasing that multiple agents can (slightly) improve accuracy rates. 
\item 
{\textbf{\color{Mulberry}Abstractive:}} 
When no additional context is provided, \texttt{gpt-3.5-turbo} achieves a mode accuracy of $53.9\%$ under perturbations. Interestingly, there is a significant dispersion of hallucination rates compared to other scenarios (Figure \ref{fig:figure-multi}). Moreover, there is significant variation in results among datasets. For instance, SQuADv2 shows a $-59.0\%$ decrease in baseline accuracy against its {\color{BrickRed} \textbf{Extractive}} counterpart. In contrast, SciQ benefits in this setting, leading to a $+9.4\%$ increase in mode accuracy, as the likelihood of generating a match increases.
\end{itemize}

\begin{figure*}[t]
\begin{lstlisting}[caption={Prompt function used for query rewriting}, language=python, label={lst:query_rewrite_prompt}]
    def prompt_creator(hallucibot_mode=None, hallucibot_outputs=None):
        # Naive Rewrite
        if hallucibot_mode is None:
            preamble = "You are a helpful expert query correction agent designed to improve user queries when needed in a manner that a novice user can easily understand. Your goal is to take an input user query and only rewrite it if you think that a novice user might find it ambigious and fail a downstream task. If it is ambigious rewrite the query while ensuring that all important information in the query given by the user is present in your rewrite. This means that any user who reads the output rewritten by you will be able to easily understand the question and answer it accurately and could cause hallucinations in a downstream task. Note that some questions might already be cleary understood by a novice user and succeed in the downstream task. These questions do not need to be rewritten."
        # HalluciBot Feedback
        elif hallucibot_mode == "basic":  
            preamble = f"You are a helpful expert query correction agent designed to improve user queries when needed in a manner that a novice user can easily understand. Your goal is to take an input user query that has been evaluated using a group of expert critics to have a majority label of `{hallucibot_outputs}` and produce a better rewritten version if needed.\
            A label of `hallucinate` is caused by ambigious text/information in a query that will cause it to be misunderstood by a novice user causing them to fail in a downstream task.\
            A label of `not hallucinate` does not have this issue and can be easily understood by a novice user who will then succeed in the downstream task. If the label is `hallucinate`, your task is to rewrite the query while ensuring that all important information in the query given by the user is present in the your rewrite. This means that any user who reads the output rewritten by you will be able to easily understand the question and answer it accurately.If the label is `not hallucinate` your task is to return the user input as is without any modifications."
        # With Consensus
        elif hallucibot_mode == "rbd":
            hb_prediction, hb_consensus_prediction = hallucibot_outputs
            if hb_consensus_prediction == "LABEL_0":
                hb_consensus_prediction = "All of the expert critics returned the same evaluation."
            elif hb_consensus_prediction == "LABEL_1":
                hb_consensus_prediction = "A minority of expert critics disagreed with the majority critic consensus about the phrase."
            preamble = f"You are a helpful expert query correction agent designed to improve user queries when needed in a manner that a novice user can easily understand. Your goal is to take an input user query that has been evaluated using a group of six expert critics to have a majority label of `{hb_prediction}` and produce a better rewritten version if needed. `{hb_consensus_prediction}`.\
            A label of `hallucinate` is caused by ambigious text/information in a query that will cause it to be misunderstood by a novice user causing them to fail in a downstream task.\
            A label of `not hallucinate` does not have this issue and can be easily understood by a novice user who will then succeed in the downstream task. If the label is `hallucinate`, your task is to rewrite the query while ensuring that all important information in the query given by the user is present in the your rewrite. This means that any user who reads the output rewritten by you will be able to easily understand the question and answer it accurately.If the label is `not hallucinate` your task is to return the user input as is without any modifications."
        
        suffix = "Carefully read the entire user query and always return the output as a JSON object with just one key labeled `rewritten_query` that corresponds to the rewritten question. DO NOT ANSWER THE GIVEN USER QUESTION AS THIS WILL AUTOMATICALLY FAIL THE DOWNSTREAM TASK."

        return preamble + "\n" + suffix
\end{lstlisting}
\end{figure*}

\subsection{Token Usage \& Statistics}\label{app:token}
We evaluated 369,837 queries, generated 1,849,185 perturbations, and generated outputs for a grand total of 2,219,022 queries. The total token usage was 717,530,842. Perturbation used 115,834,262 tokens, and the Generator used 601,696,580 tokens. Our HalluciBot is trained on 7,990,403 tokens and validated against 1,328,121 tokens, with and additional 1,305,302 tokens for testing (for \texttt{BERT-base-cased}). The lower token count is due to lack of prompts, context, and truncation for our HalluciBot models.

\subsection{HalluciBot Training Setup \& Metrics}

The parameters to our AWS EC2 training instance are provided in Table \ref{tab:aws_interface}. Additionally, we report our training metrics, including GPU hours and total floating point operations (FLOPS), in Table \ref{tab:HalluciBot-train}. Finally, the parameters for our backbones are enumerated in Table \ref{tab:HalluciBot_params}.

\subsection{Multi-class Experiments: Learning the expected value of hallucinations}

\paragraph{Multi-class Labels.} HalluciBot is trained to estimate the occurrence of hallucinations when queried and sampled under $n+1$ trials. To facilitate training, we convert the proportion into discrete classes by multiplying the original estimate $p_h(q_0)$ by the number of agents $n+1$. This transformed variable is denoted as $\mathbb{E}{\bigl[p_h(q_0)\bigr]}$.

$$\mathbb{E}{\Bigl[p_h(q_0)\Bigr]} = \Bigl \lfloor{ (n+1) \cdot p_h(q_0) }\Bigr \rfloor$$

\paragraph{Multi-class Results.}
HalluciBot achieves a validation accuracy of 47.6\%, with a Top 3 accuracy of 73.0\% for the \texttt{RoBERTa-large + Scenario} model (Table \ref{tab:HalluciBot-eval-multi}).

\begin{table*}[!t]
\centering
\begin{tabular}{l|ccc|ccc|ccc|ccc}
\hline
\multicolumn{1}{c}{} & \multicolumn{3}{c|}{\textbf{Accuracy} $\mathbf{\uparrow}$} & \multicolumn{3}{c|}{\textbf{F1 Score} $\mathbf{\uparrow}$} & \multicolumn{3}{c|}{\textbf{Precision} $\mathbf{\uparrow}$} & \multicolumn{3}{c}{\textbf{Recall} $\mathbf{\uparrow}$} \\
\cline{2-13}
\multicolumn{1}{l}{\textbf{Model}} & \textbf{Train}  & \textbf{Val}  & \textbf{Test} 
& \textbf{Train}  & \textbf{Val}  & \textbf{Test} 
& \textbf{Train}  & \textbf{Val}  & \textbf{Test}  
& \textbf{Train}  & \textbf{Val}  & \textbf{Test}  \\
\hline 
\texttt{BERT-base-cased}  &  80.9 & 64.4 & 66.5 & 81.3 & 68.6 & \underline{72.3} & 86.2 & 74.8 & 74.8 & 76.9 & 63.4 & \underline{70.0}  \\
\texttt{+ Scenario} &  \textbf{\underline{85.5}} & \underline{72.3} & 67.4  & \textbf{\underline{85.5}} & \underline{76.4} & 69.8 & \textbf{\underline{92.5}} & \underline{80.2} & \underline{77.3} & \underline{79.5} & \underline{73.0} & 63.7 \\
\hline

\texttt{RoBERTa-base}        & 74.7 & 64.1 & 66.1  &  73.3 & 66.5 & 69.6 & 85.1 & 78.0 & 74.4 & 64.4 & 57.9 & 65.3 \\

\texttt{+ Scenario}          & 79.8 & 73.0 & 69.0  &  79.3 & 76.8 & 71.7 & \underline{88.8} & \textbf{\underline{81.5}} & \textbf{\underline{78.4}} & 71.5 & 72.6 & 66.0  \\

\texttt{+ Consensus} & 79.3 & 73.0 & 68.7 & 79.1 & 77.0 & 71.5 & 87.2 & 81.0 & 77.7 & 71.4 & 73.3 & 66.2 \\

\texttt{+ Calibration} & 80.3 & \underline{\textbf{73.6}} & \textbf{\underline{69.5}} & 81.4 & 78.8 & 73.6 & 83.6 & 78.4 & 75.6 & 79.2 & 79.2 & 71.7 \\

\texttt{\quad+ $\tau=0.341$} & \underline{80.4} & \underline{\textbf{73.6}} & \underline{\textbf{69.5}} & \underline{81.6} & \textbf{\underline{80.2}} & \textbf{\underline{76.0}} & 74.7 & 72.9 & 70.3 & \underline{90.0} & \underline{89.0} & \textbf{\underline{82.6}} \\

\hline

\texttt{RoBERTa-large} & & & & & & & & & & & \\

\texttt{+ Scenario}  & \underline{84.9} & 72.9 & 68.6 & 85.0 & 76.9 & 71.1 & \underline{92.1} & \underline{80.9} & \underline{78.2} & 78.8 & 73.2 & 65.3 \\

\texttt{+ Consensus} & 83.9 & 73.1 & 68.7 & 84.0 & 77.1 & 71.7 & 90.6 & 80.8 & 77.2 & 78.3 & 73.6 & 66.9 \\

\texttt{+ Calibration} & 84.7 & 73.5 & 69.2 & \textbf{\underline{85.5}} & 78.5 & 73.0 & 88.1 & 78.9 & 76.1 & 83.1 & 78.2 & 70.1 \\

\texttt{\quad+ $\tau=0.326$} & 84.8 & \underline{\textbf{73.6}} & \underline{69.4} &83.5 & \underline{80.0} & \underline{75.6} & 75.0 & 71.8 & 70.5 & \underline{\textbf{94.2}} & \underline{\textbf{90.4}} & \underline{81.6} \\
\hline 
\end{tabular}
\caption{\label{tab:HalluciBot-eval-binary-full}
Full HalluciBot Binary Evaluation Statistics for \texttt{BERT} \citep{DBLP:journals/corr/abs-1810-04805}, \texttt{RoBERTa-base}, and \texttt{RoBERTa-large} \citep{DBLP:journals/corr/abs-1907-11692}. We report the Accuracy, F1, Precision, and Recall for all data splits. Probability threshold $\tau$ is computed along the closed interval $[0, 1]$ in increments of $0.001$ to maximize the validation F1 score for the final model. The best ablation per base model is \underline{underlined}, while the overall best performing model is in \textbf{bold}. For our experiments, the small \texttt{RoBERTa-base} outperformed the \texttt{BERT} and \texttt{RoBERTa-large} models.
}
\end{table*}

\begin{table*}[!ht]
\centering
\begin{tabular}{l|rrr|rrr|rrr}
\hline
\multicolumn{1}{c}{} &  \multicolumn{3}{c|}{\textbf{Top 1 Accuracy} $\mathbf{\uparrow}$} & \multicolumn{3}{c|}{\textbf{Top 2 Accuracy} $\mathbf{\uparrow}$} & \multicolumn{3}{c}{\textbf{Top 3 Accuracy} $\mathbf{\uparrow}$} \\ 
\cline{2-10}
\multicolumn{1}{l}{\textbf{Model}} & \textbf{Train} & \textbf{Val} & \textbf{Test}
& \textbf{Train} & \textbf{Val} & \textbf{Test}
& \textbf{Train} & \textbf{Val} & \textbf{Test} \\
\hline
\texttt{BERT-base-cased} & 49.6 & 32.2 & 24.7 & 69.7 & 49.2 & 40.7 & 81.4 & 62.7 & 56.4 \\
\texttt{+ Scenario}      & 54.1 & 38.7 & 31.3 & \underline{72.2} & \underline{54.8} & 46.1 & \underline{82.8} & \underline{67.6} & \underline{59.3} \\
\texttt{+ Ordinal}       & \underline{58.7} & \underline{45.3} & \underline{38.6} & 70.0 & 54.5 & \underline{48.3} & 79.0 & 64.1 & 59.1   \\
\hline
\texttt{RoBERTa-base} &  47.6 & 34.1 & 26.6 & 66.2 & 50.1 & 42.6 & 77.9 & 62.7 & 57.3 \\
\texttt{+ Scenario} &  \underline{52.2} & \underline{41.5} & 34.4 & \underline{69.2} & \underline{57.0} & \underline{48.4} & \underline{79.8} & \underline{68.6} & \underline{59.5}  \\
\texttt{+ Ordinal} & 47.8 & 39.4 & \underline{37.1} & 56.7 & 48.7 & 46.6 & 67.0 & 60.0 & 57.8 \\
\hline
\texttt{RoBERTa-large} & & & & & & & & &  \\
\texttt{+ Scenario} & \textbf{\underline{61.6}} & 47.6 & 38.8 & \textbf{\underline{77.5}} & \textbf{\underline{62.6}} & \textbf{\underline{53.1}} & \textbf{\underline{85.8}} & \textbf{\underline{73.0}} & \textbf{\underline{63.8}} \\
\texttt{+ Ordinal} &  60.8 & \textbf{\underline{48.0}} & \textbf{\underline{40.7}} & 73.6 & 59.0 & 52.2 & 81.9 & 67.5 & 62.3 \\
\hline

\end{tabular}
\caption{\label{tab:HalluciBot-eval-multi}
HalluciBot Multi-class Evaluation Statistics. 
Considering the challenge of approximating a random variable and the potential presence of noise in our empirical estimate, we provide accuracy measurements for Top 1, Top 2, and Top 3 predictions. 
}\vspace{3mm}
\end{table*}

\begin{table*}[!ht]
\centering
\begin{tabular}{lrrrrrrr}
\hline
\multicolumn{8}{c}{Binary (2-Class)} \\
\hline
& & GPU Time & Total & Update & Samples & Steps & Train \\
Model & Size & (Hours) & FLOP & Steps & / Second & / Second & Loss \\
\hline 
\texttt{BERT-base-cased} & 108.3M & 36.3 & 3.98E+17 & 1.89E+5 & 11.6 & 1.45 & 0.542 \\
\texttt{+ Scenario} & & 36.0 & 3.98E+17 & 1.89E+5 & 11.7 & 1.46 & 0.487 \\
\hline
\texttt{RoBERTa-base} & 124.6M &  34.5 & 3.98E+17 & 1.89E+5 & 12.2 & 1.52 & 0.574 \\
\texttt{+ Scenario} & & 36.1 & 3.98E+17 & 1.89E+5 & 11.6 & 1.46 & 0.518\\
\texttt{+ Consensus} & & 16.7 & 3.98E+17 & 1.89E+5 & 25.2 & 3.15 & 1.141 \\
\hline
\texttt{RoBERTa-large} & 355.4M \\
\texttt{+ Scenario} & & 120.5 & 1.41E+18 & 1.89E+5 & 3.5 & 0.44 & 0.495\\
\texttt{+ Consensus} & & 53.0 & 1.41E+18 & 1.89E+5 & 7.9 & 0.99 & 1.095 \\
\hline \multicolumn{8}{c}{Multi-class (7-Class)} \\
\hline
& & GPU Time & Total & Update & Samples & Steps & Train \\
Model & Size & (Hours) & FLOP & Steps & / Second & / Second & Loss \\
\hline
\texttt{BERT-base-cased} & 108.3M & 35.7 & 3.98E+17 & 1.89E+5 & 11.8 & 1.47 & 1.761 \\
\texttt{+ Scenario} & & 35.7 & 3.98E+17 & 1.89E+5 & 11.8 & 1.47 & 1.690 \\
\texttt{+ Ordinal} & & 36.8 & 3.98E+17 & 1.89E+5 & 11.4 & 1.43 & 1.778 \\
\hline
\texttt{RoBERTa-base} & 124.7M &  34.8 & 3.98E+17 & 1.89E+5 & 12.1 & 1.51 & 1.798 \\
\texttt{+ Scenario} & & 35.7 & 3.98E+17 & 1.89E+5 & 11.8 & 1.47 & 1.733 \\
\texttt{+ Ordinal} & & 34.1 & 3.98E+17 & 1.89E+5 & 12.3 & 1.54 & 1.902 \\
\hline
\texttt{RoBERTa-large} & 355.4M \\
\texttt{+ Scenario} & & 120.2  & 1.41E+18 & 1.89E+5 & 3.5 & 0.44 & 1.509\\
\texttt{+ Ordinal} & & 117.3 & 1.41E+18 & 1.89E+5 & 3.6 & 0.45 & 1.622 \\
\hline

\end{tabular}
\caption{\label{tab:HalluciBot-train}
HalluciBot training statistics for our binary and multi-class experiments. Size is the number of learnable parameters in the model. Total FLOP is the total number of floating-point operations conducted in the model. Update Steps is the number of parameter updates the Adam optimizer \citep{kingma2017adam} performs on the model. We trained each model for 5 epochs in total. Training parameters can be found in Table \ref{tab:HalluciBot_params}. There is no difference in FLOP between \texttt{BERT} and \texttt{RoBERTa} base models as \texttt{RoBERTa's} parameter increase is concentrated in a larger vocabulary. 
The \texttt{Consensus} model is fine-tuned using mixed precision (\texttt{float16}), resulting in half the training time. The loss is twice as much in the \texttt{Consensus} models given the dual loss functions for hallucination and consensus labels.
}
\end{table*}

\begin{table*}[!ht]
\centering
\begin{tabular}{l|c|c|c}
\hline
\textbf{Backbone} & \textbf{\texttt{bert-base-cased}} & \textbf{\texttt{roberta-base}}  & \textbf{\texttt{roberta-large}}\\
\hline
Transformers Version & \texttt{4.29.2} & \texttt{4.29.2} & \texttt{4.29.2} \\
Layers & 12 & 12 & 24\\
Attention Heads & 12 & 12 & 16 \\
Hidden Size & 768 & 768 & 1024\\
Intermediate Size & 3,072 & 3,072 & 4096 \\
Hidden Activation & GeLU & GeLU & GeLU \\
Hidden Dropout Prob & 0.1 & 0.1 & 0.1\\
Attention Dropout Prob & 0.1 & 0.1 & 0.1 \\
Position Embedding & Absolute & Absolute & Absolute \\
Precision &  \texttt{float32} & \texttt{float32} & \texttt{float32} \\
Max Context Length & 512 & 514 & 514 \\
Vocab Size & 28,996 & 50,265 &  50,265 \\
Total Parameters & 108.3M & 126.7M & 355.4M \\
\hline
\textbf{Learning Rate} & \textbf{\texttt{5e-6}} & \textbf{\texttt{5e-6}} & \textbf{\texttt{5e-6}}\\
Warmups & 0 & 0 & 0 \\
Scheduler Type & Linear & Linear & Linear \\
Weight Decay & 0 & 0 & 0\\
Optimizer & Adam  & Adam & Adam \\
Adam $\beta_1$ & 0.9 & 0.9 & 0.9 \\
Adam $\beta_2$ & 0.999 & 0.999 & 0.999 \\
Adam $\epsilon_1$ & 1e-8 & 1e-8 & 1e-8 \\
Max Grad Norm & 1.0  & 1.0 & 1.0 \\
\textbf{Training Batch Size} & 8 & 8 & \textbf{2} \\
\textbf{Gradient Accum. Steps} & 1  & 1 & \textbf{4} \\
\textbf{Number of Epochs} & \textbf{5} & \textbf{5} & \textbf{5} \\
\hline 
Tokenzier & \texttt{BertTokenizer} & \texttt{RobertaTokenizer} & \texttt{RobertaTokenzier}\\ 
Fast & \texttt{True} & \texttt{True} & \texttt{True} \\
\textbf{Padding Strategy} & \textbf{\texttt{max\_length}} & \textbf{\texttt{max\_length}} & \textbf{\texttt{max\_length}}  \\
\textbf{Truncation} & \textbf{\texttt{True}} & \textbf{\texttt{True}} & \textbf{\texttt{True}} \\
\textbf{Dataset Shuffle Seed} & \textbf{42} & \textbf{42} & \textbf{42} \\

\hline
\end{tabular}
\caption{\label{tab:HalluciBot_params}
Model configuration and hyper-parameters for our HalluciBots. Configurations changed from defaults highlighted in \textbf{bold}. Note that for training batch size and gradient accumulation steps, we only altered \texttt{roberta-large}'s hyper-parameters to fit onto our GPU. All models were trained using HuggingFace's Trainer class with the Adam optimizer \citep{kingma2017adam}. Note: only the \texttt{Consensus} model is tuned in \texttt{float16}, all other models are trained in full precision.
}
\end{table*}

\section{Overview of Datasets Used}
\label{app:datasets}

\begin{table*}[!ht]
\centering
\begin{tabular}{cl|lr}
\hline
\textbf{Domain}  & \textbf{Dataset}  & \textbf{Example Query} & \textbf{Answer}  \\
\hline 
Falsehoods     & TruthfulQA &
Which is denser, water vapor or air? & Air \\ \hline
Physical Env.  & PIQA &
To cut a length of PVC to size, you can? & Use a saw \\ \hline
Yes/No         & BoolQ &
Are there 51 states in the United States? & No  \\ \hline
STEM   & \multirow{3}{*}{MMLU} &
The strongest base in liquid ammonia is? & $\text{NH}_2^-$ \\
Humanities & &
When did the first pharaohs emerge in Egypt? & 5100 B.P.  \\
Social Sciences &  &
The government measures inflation using? & CPI 
\\ \hline
\multirow{4}{*}{Science} & OpenBookQA &
What raw material is consumed by chloroplast? & $\text{CO}_2$\\
& SciQ & 
Which is the final step of cell division? & Cytokinesis \\
& ARC - (C)  &
How many valence electrons does selenium have?	& $6$  \\
& ARC - (E)  &
Where is water most likely to be brackish? & Estuary \\ \hline
Mathematics & MathQA & 
If $n = 2^{0.15}$ and $n^b = 8$ , $b$ must equal? & $20$ \\ \hline
\multirow{3}{*}{Wikipedia} & SQuADv2 &
Where is the Mona Lisa housed? & The Louvre \\
& WikiQA & 
What is korean money called? & The won \\
& HotpotQA & 
EMU and Ugg boots both originated from where? & Australia \\ \hline
General & TriviaQA & 
In an opera, whose lover was Cavaradossi? & Tosca \\
\hline
\end{tabular}
\caption{\label{tab:overview}
Overview of the 13 question-answering datasets studied in this work, the domain coverage, and examples of the question-answer format. These datasets span traditional QA formats such as {\color{BrickRed} \textbf{Extractive}}, {\color{ForestGreen} \textbf{Multiple Choice}}, and {\color{Mulberry} \textbf{Abstractive}}. Our experiments treat all scenarios as a text generation problem, albeit with different prompting templates to align responses to the ground truth answer.
}
\end{table*}

\begin{table}[!t]
\centering
\small
\begin{tabular}{ll}
\hline
 \textbf{Dataset}  & \textbf{License} \\ \hline 
TruthfulQA & Apache-2.0\\ \hline
PIQA &  AFL-3.0\\ \hline
BoolQ & CC BY-SA 3.0 \\ \hline
MMLU & MIT \\ \hline
OpenBookQA & Apache-2.0\\ \hline
SciQ &  CC BY-NC 3.0\\ \hline
ARC - (C)  & CC BY-SA 4.0\\ \hline
ARC - (E)  & CC BY-SA 4.0\\ \hline
MathQA & Apache-2.0\\ \hline
SQuADv2 & CC BY-SA 4.0\\ \hline
WikiQA & Other \\ \hline
HotpotQA & CC BY-SA 4.0\\ \hline
TriviaQA & Apache-2.0\\
\hline
\end{tabular}
\vspace{1.5mm}
\caption{\label{tab:lic}
Licenses for each dataset. License for WikiQA is the Microsoft Research Data License Agreement for Microsoft Research WikiQA Corpus.
}
\end{table}

\subsection{Extractive QA}
\paragraph{SQuADv2 \citep{rajpurkar2016squad, rajpurkar2018know}} \label{squad} is the second version of the reading comprehension dataset from Stanford with 100,000 answerable and 50,000 unanswerable questions. Every answer is derived from a context $c_i$ span. For extractive QA, we only use queries that can be answered (86,821). HalluciBot is trained on 80,049 examples. For the validation set, we evaluated 5,834 items that could be answered. The reason for excluding non-answerable queries is two-fold: (1) in a zero-shot setting, it is difficult to determine if the model refuses to answer; (2) given that we are transforming the query into a semantically equivalent yet lexically distinct variation, it is impossible to determine if the new query is actually answerable from the context.

\subsection{Multiple Choice QA}
\paragraph{TruthfulQA \citep{lin-etal-2022-truthfulqa}}\label{tqa} is a QA task that gauges the ``truthfulness'' of LLMs. 817 questions encompass 38 categories intended to elicit false beliefs in domains such as ``misconceptions'' or ``conspiracy theories''. 
For the multiple choice approach, we provide all candidates with only one being correct. The Generator will then select.  TruthfulQA helps measure whether our perturbed variations can act as an adversarial force to elicit false beliefs. 

\paragraph{PIQA \citep{Bisk2020}} is the ``Physical Interaction: Question Answering'' dataset focusing on physical commonsense through dichotomous choices. 
PIQA tests the ability of LLMs to understand how to use everyday materials; it consists of 16,113 training and 1,838 validation samples. 

\paragraph{MMLU \citep{hendryckstest2021, hendrycks2021ethics}} features 57 subjects from science, technology, law, humanities, and social sciences in a multiple choice format. In this experiment, each query fed to the Output Generator will be provided the perturbed query and the original answer choices. Then, it will be asked for the best choice. Accuracy is measured by the best match of the label/choice in the set of responses. There are 14,042 test, 1,531 validation, and 285 development samples.

\paragraph{OpenBookQA \citep{OpenBookQA2018}} tests scientific commonsense knowledge comprised of elementary science multiple choice questions from the WorldTree corpus. The train-val-test split is 4,957 training, 500 validation, and 500 testing samples. 

\paragraph{BoolQ \citep{clark2019boolq, wang2019superglue}} is a dichotomous QA set of yes/no questions. 
In our study, we exclude context and only ask for Yes/No or True/False answers. There are 9,427 training, 3,270 validation, and 3,245 test samples.

\paragraph{SciQ \citep{SciQ}}\label{sciq} contains 13,679 science questions covering Physics, Biology, and Chemistry, broken into 11,679 training, 1,000 validation, and 1,000 test samples. Each query is paired with an answer and three distractors. We omit any supporting evidence and rely solely on candidates and the Generator's general knowledge. Since the multiple-choice answer is always consistently situated in the order of choices, we randomize the answers once before formatting.  This is to enforce the model to rely on semantics instead of ordinal patterns. 

\paragraph{ARC - Challenge \citep{allenai:arc}} is the AI2 Reasoning Challenge set consisting of 7,787 questions from grade-school level science exams. The Challenge set contains 2,590 \textit{hard} questions, amalgamated from samples that were answered incorrectly by both retrieval and word co-occurrence algorithms. There are 1,119 training, 299 validation, and 1,172 test examples. Most samples have 4 answer choices, with $<1\%$ having 3 or 5 choices.

\paragraph{ARC - Easy \citep{allenai:arc}} is the AI2 Reasoning Easy set.  
It consists of 5,197 questions with a train-val-test split of 2,251, 570, and 2,376.

\paragraph{MathQA \citep{amini-etal-2019-mathqa}} allows us to isolate our system on mathematical problems. The test set has no answers; therefore, our evaluations focus on the validation set of 4,475 samples.

\subsection{Abstractive QA}
\paragraph{SQuADv2 \citep{rajpurkar2016squad, rajpurkar2018know}}
is a reading comprehension taskset (\S\ref{squad}) that we repurposed for abstractive QA. By omitting the context from the Generator's prompt, the model is conditioned on the transformed query alone. We juxtapose the differences in metrics between the extractive and abstractive settings in Table \ref{results}. 

\paragraph{TruthfulQA \citep{lin-etal-2022-truthfulqa}}
as referenced in \S\ref{tqa}, has 817 questions to reveal misconceptions. For abstractive QA, the LLM can construct any free-text answer and match the result against the candidates via cosine similarity. 
The best match is either correct or a distractor. 

\paragraph{WikiQA \citep{yang-etal-2015-wikiqa}}
comprises of 3,047 questions and 29,258 sentence pairings. Although originally crafted for information retrieval tasks, we repurposed WikiQA as an abstractive task by filtering for the 1,473 sentences that were labeled as the answer to the corresponding questions. From these QA pairings, we generate answers for each perturbation. Then the labeled passage and generated sentence are aligned. If there is a high semantic similarity $(> 60\%)$ between the answer and passage, the answer is considered correct. 

\paragraph{SciQ \citep{SciQ}} as mentioned in \S\ref{sciq}, evaluates 13,679 science exam questions \textit{without} candidate choices. Correct responses are measured by the approximate Levenshtein distance of label substrings in the generated answer. 

\paragraph{HotpotQA \citep{yang-etal-2018-hotpotqa, petroni-etal-2021-kilt}} consists of over 113,000 Wikipedia-based QA pairs. This set provides short, factual answers and is topically diverse. We evaluate 57,711 train samples and 5,600 validation samples. The test set contains no answers; therefore, we cannot discern any meaningful metrics.

\paragraph{TriviaQA \citep{2017arXivtriviaqa}}
diversifies perturbations to general knowledge domains, leveraging 95,000 syntactically and lexically variable QA pairs authored by trivia enthusiasts. 
Only the training and validation splits contain answers; correspondingly, we evaluate 11,313 validation and 67,469 training samples. 

\section{Metrics} \label{sec:app_metrics}
We outline our metrics used in assessing the Multi-Agent Monte Carlo simulation. We also bifurcate whether metric requires labels to compute, denoted as \textbf{(S)}upervised and \textbf{(U)}nsupervised metrics.

\subsection{Notation}
\[
\begin{array}{rl}
q_{j,0} & \text{$j$-th example, original query}\\
q_{j,i} & \text{$j$-th example, $i$-th perturbation}\\
\mathcal{Q}_j & \text{Set of query variations for $q_{j,i}\,\forall i$}\\
y_{j} & \text{Ground truth for query $q_{j,0}$}\\
a_{j,i} & \text{$j$-th example, $i$-th output}\\
\mathcal{A}_j & \text{Set of outputs for $q_{j,i}\,\forall i$}\\
n & \text{Number of perturbations}\\
n + 1 & \text{Number of perturbation with $q_{j,0}$}\\
m & \text{Number of examples} \\ 
k_j & \text{Number of allowed answer states} \\
f_{j,i} & \text{Frequency of $a_{j,i}$ across all $i$}
\end{array}
\]

\subsection{Accuracy (S)}\label{robust}
Since the original query $q_{j,0}$ is in our answer set $\mathcal{A}_j$, we can juxtapose the baseline accuracy against ensemble metrics, such as lower-bound performance, upper-bound performance, and plurality-based accuracy. 
Inspired by prior work evaluating LLMs \citep{liang2022holistic}, we measure robustness in the worst-case (one or more raters is incorrect) and best-case (one rater is correct) scenarios. If we let $\mathbb{I}$ be the indicator function for partial matches, then let $A$ be the baseline accuracy for $n$ samples. In addition, let $\Omega$ be the lower bound, i.e. worst-case performance under perturbations $\mathcal{Q}_j$, and let $O$ be the upper-bound, i.e. best-case performance. Finally, given the set of $n+1$ raters, let $\hat{Y}$ be the aggregate responses by the mode of answer set $\mathcal{A}_j$, as a proxy for plurality voting. 
\begin{gather*}
    A = \frac{1}{m}\sum_{j=1}^{m} \mathbb{I}\Bigl[a_{j,0} = y_j\Bigr] \\ 
    \Omega = \frac{1}{m}\sum_{j=1}^{m}\min_i \mathbb{I}\Bigl[a_{j,i} = y_j\Bigr] \leq A \\
    O = \frac{1}{m}\sum_{j=1}^{m}\max_i \mathbb{I}\Bigl[a_{j,i} = y_j\Bigr] \geq A \\
    \hat{Y} = \frac{1}{m}\sum_{j=1}^{m} \mathbb{I}\Bigl[ \text{mode} \bigl\{\, a_{j,i} \,\bigm| a_{j,i} \in \mathcal{A}_j\, \bigr\} = y_j \Bigr] 
\end{gather*}
Since $q_{j,0}$ is the original query, the relationship between accuracy $A$, lower-bound performance $\Omega$, and upper-bound performance $O$ is as follows: $$\Omega \leq A \leq O$$
If our raters randomly guessed, then accuracy $A$ and mode $\hat{Y}$ for $k_j$ choices would be $1/k_j$. The lower-bound $\Omega$ would approach: $$\Omega = \lim_{k_j\to \infty}\left(\frac{1}{k_j}\right)^{n+1} \approx 0$$ 
The upper-bound performance $O$ would then be the probability of one success for $n+1$ trials. 
$$O = 1 - \left(\frac{k_j-1}{k_j}\right)^{(n+1)}$$

\subsection{Agreement}\label{aggreement}

\subsubsection{Item Difficulty (S)}
For each split, the average item difficulty $\mu_D$ is the mean percentage of correct responses per query \citep{Lord1952}. It is a measure of the collective difficulty of the queries for our LLM raters. The baseline for random guessing is the expected value of a Bernoulli distribution, $\mathbb{E}\left[\mu_D\right] = 1 / k$. 
\begin{gather*}
    \mu_D = \frac{1}{m}\sum_{j=1}^{m}  \Biggl( \frac{1}{n+1}\sum_{i=0}^{n} 
    \mathbb{I}\Bigl[a_{j,i} = y_j\Bigr] \Biggr) 
\end{gather*}

\subsubsection{Mean Normalized Certainty (U)}
Entropy $H$ can quantify the degree of uncertainty in a set of qualitative responses. It is maximized for uniform distributions (complete uncertainty) while minimized for consistent categorizations \citep{6773024, 4d25ef96-6507-346e-8e18-720c9de36b78}. 
We normalize the rater entropy $H$ by the maximum entropy allowed $H_{max}$. We reverse the scale such that 1 indicates certainty, 0 for uncertainty. Let $f_{j,i}$ denote the frequency of answer candidate $a_i$ for example $q_j$, and $k_j$ be the number of total allowable choices (unique states). Then proportion $p_{j,i}$ and mean normalized certainty (MNC) $H_{\eta}$ is:
\begin{align*}
 p_{j,i} &= \frac{f_{j,i}}{n+1} \\
H_{\eta} &= 1 - \mathbb{E} \left[\frac{H}{H_{max}}\right] \\
&= 1 + \frac{1}{m}\sum_{j=1}^{m} \left[\sum_{i=0}^{k_j} \frac{p_{j,i}  \log \left(p_{j,i}\right)}{\log \left(k_j\right)} \right]
\end{align*}

\subsubsection{Gibbs' M2 Index (U)}
This index is a standardized metric measuring the ratio of the variance of a multinomial distribution to the variance of a binomial; since each perturbation is an independent trial, and the answer responses are categorized into exactly one of $k_j$ outcomes, each round of our Monte Carlo sampling is a multinomial simulation \citep{10.1093/sf/53.3.468}. Therefore, let $p_{j,i}$ be the proportion of observations for the $i$-th category and $k_j$ be the number of allowed categories. For readability, we reverse the index such that $M_2=1$ when our raters are certain and 0 when uniform.
\begin{gather*}
    M_2 = 1 - \frac{1}{m}\sum_{j=1}^{m}\left[\frac{k_j}{k_j - 1} \left(
1 - \sum_{i=0}^{k_j} \Bigl(p_{j,i}\Bigr)^2
\right)\right]
\end{gather*}

\subsubsection{Fleiss's Kappa (U)}
Inter-rater agreement is measured through Fleiss' Generalized $\kappa$ \citep{doi:10.1177/001316446002000104, Fleiss1971}. 
This metric calculates the degree of agreement in responses over what would be expected by random chance, $1$ indicating complete agreement (and 0 for none). 
Let $f_{j,i}$ be the frequency of answer choice $a_i$ for example $q_j$, then the expected agreement by chance $\bar{P_e}$ and observed agreement $\bar{P_o}$ for $n+1$ raters is
\begin{gather*}
    \bar{P}_e = \sum_{i=0}^{k_j}\left(\frac{1}{m(n+1)}\sum_{j=1}^{m} f_{j,i} \right)^2 \\
    P_j = \frac{1}{n(n+1)}
        \left[
        \left(\sum_{i=0}^{k_j} \Bigl(f_{j,i}\Bigr)^2\right) - (n+1)
        \right] \\
    \bar{P}_o = \frac{1}{m} \sum_{j=1}^{m} P_j
\end{gather*}
Then $\kappa$ is the ratio of the degree of agreement achieved over the degree of agreement attainable through pure chance. Note that $\kappa$ is affected by the number of raters and categories, with fewer categories often yielding higher $\kappa$ values.
\begin{gather*}\kappa = \frac{\bar{P_o} - \bar{P_e}}{1-\bar{P_e}}\end{gather*}

\subsection{Reliability: Cronbach's Alpha (S)}
For measures of internal consistency, we rely on Cronbach's $\alpha$ \citep{Cronbach1951} for dichotomous choices, in which 1 is for correct and 0 is for incorrect. We choose Cronbach's $\alpha$ as it is widely accepted in testing theory, and is equivalent to the Kuder-Richardson Formula 20 (KR-20) for binary data \citep{RePEc:spr:psycho:v:2:y:1937:i:3:p:151-160}. Let $m$ be the number of samples, $\sigma^2_y$ be each sample's score variance across our $n+1$ raters, and $\sigma_x^2$ be the variance across the total count of correct responses per rater. Then Chronbach's $\alpha$ is defined as: 
\begin{gather*}
\alpha = \frac{m}{m-1} \left( 1 - \frac{\sum_{j=1}^{m} \sigma^2_y}{\sigma^2_x} \right)
\end{gather*}

\section{Abusive or Sensitive Content}
Throughout our experiments, when able, we captured statistics surrounding \texttt{gpt-3.5-turbo}'s failure to provide any response at all. While a vast majority were related to latency and servicing (\texttt{APIError}, \texttt{ServiceUnavailableError}, \texttt{RateLimitError}), a small subset of 2,293 samples registered \texttt{AttributeError} (1,081) or \texttt{InvalidRequestError} (1,212). The former is triggered when we generate violent or explicit content. The latter is triggered through prompt filtering, such as violent or explicit terms appearing in the prompt. In either case, we enumerate the split per dataset as follows:
\[
\begin{array}{rlrl}
\text{MMLU} & 638 &
\text{PIQA} & 543\\
\text{SQuADv2} & 490 &
\text{TriviaQA} & 326 \\
 \text{HotpotQA} & 103 &
 \text{BoolQ} & 57  \\
 \text{OpenBookQA} & 48 &
 \text{TruthfulQA} & 29\\
 \text{SciQ} & 25  &
 \text{WikiQA} & 19 \\
 \text{MathQA} & 15 &
 \text{Total} & 2,293
\end{array}
\]

\section{Sample Perturbations \& Quality}
\label{sec:appendix}\label{sec:wellformresults}
We showcase 10 examples across our scenarios, alongside predicted hallucination rate from our Accuracy-Agreement tuned model, in Table \ref{tab:psamples}.
We used a syntactically-aware well-formedness scoring \texttt{RoBERTa} model \citep{WinNT} trained on the Google Query Well-formedness Dataset \citep{faruqui-das-2018-identifying} to evaluate the grammatical correctness and completeness of 1,881,005 synthetically generated queries. 
We present our well-formedness results across each scenario and dataset in Table \ref{tab:wellformedness}.

Our analysis indicates that the perturbations created by \texttt{gpt-3.5-turbo} consistently exhibit a high level of coherence, as indicated by their well-formedness score of 0.87. In contrast, the original queries achieve a well-formedness score of 0.77, representing an 11.5\% decline. Table \ref{tab:wellformedness} expands on our results, with sample perturbations in Table \ref{tab:psamples}.

\begin{table}[!t]
\centering
\begin{tabular}{llrr}
\hline
 & Scope &  Original &   Generated    \\
 \hline\hline
\multirow{1}{*}{\rotatebox[origin=c]{90}{{\color{BrickRed} \textbf{Ex}}}}
& SQuADv2 &      0.86 & 0.91 \\
\hline
 \multirow{9}{*}{\rotatebox[origin=c]{90}{{\color{ForestGreen} \textbf{Multiple Choice}}}}  
                  & TruthfulQA &      0.91 & 0.92 \\
                  & PIQA &      0.59 & 0.96 \\
                  & MMLU &      0.63 & 0.83 \\
                  & OpenBookQA &      0.56 & 0.90 \\
                  & BoolQ &      0.78 & 0.95 \\
                  & SciQ &      0.79 & 0.91 \\
                  & ARC - Challenge &      0.80 & 0.88 \\
                  & ARC - Easy &      0.81 & 0.89 \\
                  & MathQA &      0.33 & 0.75 \\
                  \hline
\multirow{6}{*}{\rotatebox[origin=c]{90}{{\color{Mulberry} \textbf{Abstractive}}}} 
                  & SQuADv2 &      0.86 & 0.91 \\
                  & TruthfulQA &      0.91 & 0.93 \\
                  & WikiQA &      0.71 & 0.91 \\
                  & SciQ &      0.79 & 0.91 \\
                  & HotpotQA (KILT) &      0.72 & 0.81 \\
                  & TriviaQA &      0.80 & 0.85 \\
\hline \hline
\multirow{2}{*}{\rotatebox[origin=c]{90}{{\color{BrickRed} \textbf{Ext}}}} 
                  & train &      0.86 & 0.91 \\
                  & validation &      0.85 & 0.91 \\
                  \hline
 \multirow{3}{*}{\rotatebox[origin=c]{90}{{\color{ForestGreen} \textbf{MC}}}}  
                  & train &      0.69 & 0.93 \\
                  & validation &      0.60 & 0.86 \\
                  & test &      0.65 & 0.85 \\
    \hline
\multirow{3}{*}{\rotatebox[origin=c]{90}{{\color{Mulberry} \textbf{Abstr}}}} 
                  & train &      0.78 & 0.85 \\
                  & validation &      0.80 & 0.86 \\
                  & test &      0.76 & 0.91 \\
\hline \hline
\multirow{6}{*}{\rotatebox[origin=c]{90}{{\color{WildStrawberry} \textbf{Totals}}}} 
& train      &      0.79 & 0.88 \\
& validation &      0.75 & 0.87 \\
& test       &      0.65 & 0.85 \\
\cline{2-4} 
& {\color{BrickRed} \textbf{Extractive} }     &      0.86 & 0.91 \\
& {\color{ForestGreen} \textbf{Multiple Choice}} &      0.66 & 0.89 \\
& {\color{Mulberry} \textbf{Abstractive}}    &      0.78 & 0.85 \\
\hline \hline
& \textbf{Aggregate Total} &   0.77 &  0.87 \\
\hline
\end{tabular}\vspace{2mm}
\caption{\label{tab:wellformedness}
Well-formedness scores of the original datasets juxtaposed to the average well-formedness score of \texttt{gpt-3.5-turbo} generated queries, with each sample averaged across the $n$ perturbed queries. We do not split our datasets between training splits as there was no significant different in scores. We color {\color{BrickRed} \textbf{Extractive}}, {\color{ForestGreen} \textbf{Multiple Choice}}, and {\color{Mulberry} \textbf{Abstractive}} scopes appropriately to differentiate scenarios, training splits, and totals.
}
\end{table}

\begin{table*}[!th]
\centering
\small
\begin{tabular}{c|clr}
\hline
 & $\mathbf{Q_i}$ & \textbf{Query} & \textbf{\%}  \\
\hline
\multirow{6}{*}{\rotatebox[origin=c]{90}{{\color{BrickRed} \textbf{SQuADv2}}}} 
& $q_0$ & How did Frederick protect Silesia when he went to invade Saxony? & \gradientcolor{44.2}  \\
& $q_1$ & What measures did Frederick take to ensure the safety of Silesia during his invasion of Saxony? & \gradientcolor{25.8} \\
& $q_2$ & In what ways did Frederick safeguard Silesia while he was engaged in conquering Saxony? & \gradientcolor{87.3} \\
& $q_3$ & During his invasion of Saxony, how did Frederick ensure the security of Silesia? & \gradientcolor{41.8} \\
& $q_4$ & What steps did Frederick take to defend Silesia while simultaneously attacking Saxony? & \gradientcolor{64.0}  \\
& $q_5$ & How did Frederick maintain the protection of Silesia while leading his troops into Saxony? & \gradientcolor{64.0}  \\
\hline

\multirow{6}{*}{\rotatebox[origin=c]{90}{{\color{ForestGreen} \textbf{PIQA}}}}
& $q_0$ & How do you cut up a watermelon? & \gradientcolor{15.6} \\
& $q_1$ & What is the proper way to slice a watermelon? & \gradientcolor{18.5} \\
& $q_2$ & Can you explain the process of cutting a watermelon into small pieces? & \gradientcolor{9.8}  \\
& $q_3$ & Is there a certain technique to chop a watermelon effectively? & \gradientcolor{17.7}  \\
& $q_4$ & What are the steps involved in cutting a watermelon? & \gradientcolor{21.8}  \\
& $q_5$ & Could you share some tips on how to cut a watermelon into bite-sized pieces? & \gradientcolor{29.6} \\
\hline

\multirow{6}{*}{\rotatebox[origin=c]{90}{{\color{ForestGreen} \textbf{BoolQ}}}}
& $q_0$ &  Is there compulsory military service in the US? & \gradientcolor{18.7}\\
& $q_1$ &  Does the US mandate military service for its citizens? & \gradientcolor{12.5}\\
& $q_2$ &  Is military conscription mandatory in the United States? & \gradientcolor{12.3}\\
& $q_3$ &  Do US citizens have to serve in the military? & \gradientcolor{73.7}\\
& $q_4$ &  Is it required for individuals in the US to participate in the armed forces? & \gradientcolor{39.7}\\
& $q_5$ &  Is there a legal obligation for Americans to serve in the military? & \gradientcolor{18.2}\\
\hline

\multirow{6}{*}{\rotatebox[origin=c]{90}{{\color{ForestGreen} \textbf{TruthfulQA}}}}
& $q_0$ & What did the word "posh" originally stand for?  & \gradientcolor{7.7} \\
& $q_1$ & What was the original meaning of the term "posh"? & \gradientcolor{18.0} \\
& $q_2$ & From where did the word "posh" get its start?  & \gradientcolor{31.1} \\
& $q_3$ & Can you enlighten me about the etymology of the word "posh"?  & \gradientcolor{14.8} \\
& $q_4$ & What was the initial intention of the term "posh"?  & \gradientcolor{60.6} \\
& $q_5$ & What did "posh" signify when it first came into use?  & \gradientcolor{7.3} \\
\hline

\multirow{6}{*}{\rotatebox[origin=c]{90}{{\color{ForestGreen} \textbf{SciQ}}}}
& $q_0$ & What are found in moist forests that break down decaying plant material?  & \gradientcolor{6.7}  \\
& $q_1$ & Which organisms decompose decaying plant material in damp forests?  & \gradientcolor{23.2}  \\
& $q_2$ & Name the species present in wet forests that aid in the breakdown of decaying plant matter?  & \gradientcolor{5.7}  \\
& $q_3$ & What living beings inhabit moist forests and are responsible for the decomposition of decaying plant material? & \gradientcolor{6.7}   \\
& $q_4$ & In what type of forests can we find organisms that decompose rotting plant material?  & \gradientcolor{9.2}  \\
& $q_5$ & Which creatures are responsible for breaking down decomposing plant matter in damp woodland areas?  & \gradientcolor{11.0}  \\
\hline

\multirow{6}{*}{\rotatebox[origin=c]{90}{{\color{ForestGreen} \textbf{ARC - C}}}}
& $q_0$ & Which biomolecule does not have a carbon-nitrogen bond? &  \gradientcolor{21.1}\\
& $q_1$ & Among all biomolecules, which one lacks a bond between carbon and nitrogen atoms? &  \gradientcolor{15.5}\\
& $q_2$ & Which of the biomolecules do not contain a carbon-nitrogen linkage? &  \gradientcolor{37.9}\\
& $q_3$ & Can you name the biomolecule which does not exhibit a bond between nitrogen and carbon atoms? &  \gradientcolor{6.1}\\
& $q_4$ & What is the biomolecule which doesn't have any carbon-nitrogen bonds? &  \gradientcolor{6.3}\\
& $q_5$ & Identify the biomolecule that doesn't have a bond between nitrogen and carbon. &  \gradientcolor{22.6}\\
\hline

\multirow{6}{*}{\rotatebox[origin=c]{90}{{\color{ForestGreen} \textbf{MMLU}}}}
& $q_0$ & A writ of certiorari from the Supreme Court indicates that the Court & \gradientcolor{20.6} \\
& $q_1$ & The Supreme Court has issued a writ of certiorari, what does this signify? & \gradientcolor{20.2} \\
& $q_2$ & What is the implication of the Supreme Court issuing a writ of certiorari? & \gradientcolor{79.3} \\
& $q_3$ & The Supreme Court has granted a writ of certiorari, what does this mean? & \gradientcolor{37.0} \\
& $q_4$ & What is the significance of the Supreme Court granting a writ of certiorari? & \gradientcolor{73.1} \\
& $q_5$ & What does it mean when the Supreme Court issues a writ of certiorari? & \gradientcolor{17.8} \\
\hline

\multirow{6}{*}{\rotatebox[origin=c]{90}{{\color{Mulberry} \textbf{WikiQA}}}}
& $q_0$ & How was color introduced in film?  & \gradientcolor{85.4} \\
& $q_1$ & What is the history of incorporating color in movies?  & \gradientcolor{93.9} \\
& $q_2$ & How did the implementation of color in films come about?  & \gradientcolor{98.3} \\
& $q_3$ & What was the process behind introducing color into motion pictures?  & \gradientcolor{90.2} \\
& $q_4$ & When and how did filmmakers start using color in their productions?  & \gradientcolor{96.2} \\
& $q_5$ & What is the story behind the integration of color into the film industry? & \gradientcolor{98.8}  \\
\hline

\multirow{6}{*}{\rotatebox[origin=c]{90}{{\color{Mulberry} \textbf{HotpotQA}}}}
& $q_0$ & What state was the man that Atchison County was named after from?  & \gradientcolor{51.3} \\
& $q_1$ & From which state did the person who gave the name Atchison County hail?  & \gradientcolor{70.1} \\
& $q_2$ & What was the home state of the individual after whom Atchison County was named?  & \gradientcolor{84.6} \\
& $q_3$ & Which state did the namesake of Atchison County belong to? & \gradientcolor{6.1}  \\
& $q_4$ & What state did the person who inspired the name of Atchison County belong to? & \gradientcolor{42.3}  \\
& $q_5$ & To which state did the man after whom Atchison County was named originally belong? & \gradientcolor{67.3}  \\
\hline

\multirow{6}{*}{\rotatebox[origin=c]{90}{{\color{Mulberry} \textbf{TriviaQA}}}}
& $q_0$ & Which English king ruled for the shortest period? & \gradientcolor{82.9}  \\
& $q_1$ & Who is the English king with the briefest reign? & \gradientcolor{82.8}  \\
& $q_2$ & Which king of England had the shortest time in power? & \gradientcolor{83.4}  \\
& $q_3$ & Can you name the English monarch who had the quickest reign? & \gradientcolor{72.5}  \\
& $q_4$ & Which royal ruler of England had the shortest reign length? & \gradientcolor{92.8}  \\
& $q_5$ & What was the name of the king of England with the shortest reign period? & \gradientcolor{92.2}  \\
\hline

\end{tabular}
\caption{\label{tab:psamples} Sample perturbations split by dataset, colored by task scenario. The Query Perturbator produces lexically distinct variations while retaining key semantic information. However, as our experiments show, variations can predispose \texttt{gpt-3.5-turbo} to hallucinations. Recall that $q_0$ is the original, unaltered query. We include HalluciBot's predicted hallucination probability for the positive class (``Yes'' to observe at least one hallucination).
}
\end{table*}

\begin{table*}[!th]
\centering
\begin{tabular}{cllr|cc|cc|cccc|c}
\hline
\multicolumn{4}{c|}{\textbf{Datasets}} & \multicolumn{4}{c|}{\textbf{Accuracy}} & \multicolumn{4}{c|}{\textbf{Agreement}} & \multicolumn{1}{c}{\textbf{Rel}}\\
\hline
& \textbf{Name} & \textbf{Split} & \textbf{\#} & \multicolumn{1}{c}{\textbf{Base} $\mathbf{\uparrow}$} & \textbf{Mode} $\mathbf{\uparrow}$ & \textbf{Lower} $\mathbf{\uparrow}$ & \textbf{Upper} $\mathbf{\uparrow}$ & $\mathbf{\mu_D}$ $\mathbf{\uparrow}$ & $\mathbf{H_{\eta}}$ $\mathbf{\uparrow}$ & $\mathbf{M_2}$ $\mathbf{\uparrow}$ & $\mathbf{\kappa}$ $\mathbf{\uparrow}$ & $\mathbf{\alpha}$ $\mathbf{\uparrow}$ \\
\hline\hline
\multirow{2}{*}{\rotatebox[origin=c]{90}{{\color{BrickRed} \textbf{Extn}}}} &
 \multirow{2}{*}{SQuADv2} & train & 80,049 & \textbf{91.9} & 90.8 & 68.6 & 97.3 & 87.0 & 85.8 & 84.4 & 75.0 & 99.9 \\
     &    & val & 5,843 & \textbf{95.2} & 94.1 & 74.5 & 98.8 & 90.6 & 81.2 & 82.7 & 79.3 & 98.3 \\
\hline\hline
 \multirow{23}{*}{\rotatebox[origin=c]{90}{{\color{ForestGreen} \textbf{Multiple Choice}}}} &
 TruthfulQA & val & 786 & \textbf{60.4} & 60.3 & 39.8 & 76.1 & 58.5 & 88.1 & 79.5 & 72.6 & 37.8 \\
 \cline{2-13}
 &\multirow{2}{*}{PIQA} & train & 15,677 & 81.2 & \textbf{82.3} & 56.7 & 94.0 & 78.8 & 79.1 & 77.6 & 65.1 & 96.8  \\
 &     & val & 1,784 & 80.1 & \textbf{83.2} & 58.2 & 93.8 & 79.2 & 79.3 & 77.9 & 66.0 & 82.7 \\
\cline{2-13}
 & \multirow{3}{*}{MMLU} & dev & 281 & \textbf{66.2} & 61.9 & 35.6 & 82.6 & 58.9 & 74.8 & 68.7 & 60.6 & 74.4 \\
&  & val & 1,463 & 64.5 & \textbf{65.0} & 37.9 & 82.8 & 60.3 & 74.8 & 68.8 & 60.8 & 85.9 \\
&  & test & 13,545 & \textbf{67.6} & 67.3 & 38.4 & 84.2 & 61.6 & 75.0 & 69.0 & 61.1 & 99.1  \\
 \cline{2-13}
 & \multirow{3}{*}{\begin{tabular}{@{}l} OpenBook \\ QA \end{tabular}} & train & 4,909 & \textbf{78.0} & 75.6 & 37.8 & 91.3 & 67.8 & 72.8 & 66.6 & 58.0 & 99.1 \\
 & & val & 497 & 78.1 & \textbf{78.9} & 39.8 & 93.0 & 70.5 & 73.4 & 67.5 & 59.2 & 87.1  \\
 & & test & 499 & \textbf{75.6} & 73.7 & 38.1 & 90.8 & 66.8 & 73.1 & 66.9 & 58.4 & 88.9 \\
 \cline{2-13}
 & \multirow{2}{*}{BoolQ} & train & 9,401 & 71.0 & \textbf{71.2} & 32.7 & 92.6 & 67.0 & 51.2 & 54.3 & 43.1 & 97.0 \\
 & & val & 3,256 & 71.5 & \textbf{71.7} & 33.5 & 93.2 & 67.6 & 51.5 & 54.7 & 43.3 & 91.6 \\
 \cline{2-13}
 &\multirow{3}{*}{SciQ} & train & 11,670 & \textbf{93.4} & 92.9 & 76.7 & 97.6 & 89.9 & 91.7 & 89.4 & 86.4 & 98.7 \\
 & & val & 999 & 91.6 & \textbf{93.0} & 77.3 & 97.3 & 89.9 & 91.7 & 89.6 & 86.7 & 45.5  \\
 & & test & 998 & \textbf{93.8} & 93.5 & 76.9 & 97.8 & 90.6 & 91.9 & 89.8 & 87.0 & 85.6 \\
 \cline{2-13}
 &\multirow{3}{*}{\begin{tabular}{@{}l} ARC - \\ Challenge \end{tabular}} & train & 1,118 & \textbf{85.5} & 82.9 & 52.3 & 95.2 & 76.8 & 82.7 & 76.9 & 69.9 & 95.6 \\
 & & val & 299 & \textbf{87.0} & 82.3 & 53.8 & 95.0 & 77.1 & 82.5 & 76.4 & 68.9 & 88.9 \\
 & & test & 1,172 & \textbf{83.8} & 80.7 & 50.9 & 92.7 & 74.7 & 81.1 & 76.3 & 70.3 & 96.1 \\
 \cline{2-13}
 &\multirow{3}{*}{\begin{tabular}{@{}l} ARC - \\ Easy \end{tabular}} & train & 2,248 & \textbf{93.3} & 92.7 & 70.8 & 97.9 & 88.0 & 90.1 & 87.4 & 83.7 & 96.6  \\
 & & val & 570 & \textbf{94.4} & 92.1 & 66.1 & 98.1 & 86.7 & 87.6 & 84.6 & 80.8 & 92.4  \\
 & & test & 2,374 & \textbf{92.8} & 92.5 & 69.6 & 98.3 & 87.9 & 90.2 & 86.8 & 82.9 & 95.7  \\
 \cline{2-13}
 &\multirow{3}{*}{MathQA} & train* & 693 & 50.1 & \textbf{56.9} & 9.5 & 85.9 & 46.6 & 60.1 & 46.1 & 30.2 & 39.0 \\
 & & val & 4,473 & 49.8 & \textbf{55.5} & 9.4 & 85.8 & 45.9 & 64.4 & 47.8 & 29.8 & 89.7  \\
 & & test & 2,985 & 47.7 & \textbf{54.7} & 9.2 & 84.5 & 45.6 & 68.1 & 50.0 & 31.3 & 45.2 \\
\hline\hline
\multirow{13}{*}{\rotatebox[origin=c]{90}{{\color{Mulberry} \textbf{Abstractive}}}}
 & \multirow{2}{*}{SQuADv2} & train & 20,842 & \textbf{32.9} & 31.8 & 15.1 & 46.7 & 29.9 & 74.7 & 76.4 & 66.3 & 98.5  \\
   &      & val & 5,864 & \textbf{25.4} & 24.0 & 10.2 & 37.9 & 22.9 & 90.4 & 86.3 & 64.6 & 94.3  \\
 \cline{2-13}
 & TruthfulQA & val & 807 & \textbf{52.4} & 28.1 & 61.8 & 78.6 & 55.1 & 58.9 & 61.5 & 53.4 & 31.3  \\
 \cline{2-13}
 & \multirow{3}{*}{WikiQA} & train & 1,028 & \textbf{73.4} & 72.6 & 54.6 & 80.9 & 69.8 & 79.6 & 81.3 & 77.6 & 86.5  \\
 & & val & 140 & \textbf{76.4} & 75.7 & 58.6 & 82.9 & 73.3 & 80.9 & 82.4 & 78.3 & 94.2 \\
 & & test & 286 & \textbf{73.8} & 69.9 & 52.4 & 81.1 & 67.8 & 76.8 & 78.4 & 74.0 & 80.6  \\

 \cline{2-13}
 & \multirow{3}{*}{SciQ} & train & 11,596 & 66.2 & \textbf{75.6} & 35.3 & 80.4 & 59.8 & 80.6 & 74.7 & 69.0 & 99.2 \\
 & & val & 991 & 65.7 & \textbf{74.7} & 34.8 & 80.1 & 58.9 & 80.6 & 74.8 & 68.9 & 91.0 \\
 & & test & 995 & 70.1 & \textbf{77.4} & 36.8 & 83.2 & 62.4 & 80.5 & 74.6 & 69.1 & 93.5 \\
 \cline{2-13}
 & \multirow{2}{*}{\begin{tabular}{@{}l} HotpotQA \\ (KILT) \end{tabular}} & train & 66,345 & \textbf{45.5} & 41.7 & 21.2 & 59.6 & 40.6 & 80.7 & 78.6 & 65.8 & 99.8 \\
 & & val & 5,542 & \textbf{42.5} & 38.8 & 21.5 & 55.9 & 38.3 & 72.5 & 74.4 & 69.3 & 96.8  \\
 \cline{2-13}
& \multirow{2}{*}{TriviaQA} & train & 76,635 & \textbf{71.7} & 69.5 & 48.6 & 79.7 & 66.8 & 84.6 & 83.1 & 72.8 & 99.9  \\
 & & dev & 11,177 & \textbf{72.2} & 69.8 & 48.5 & 80.1 & 67.0 & 84.8 & 82.8 & 72.5 & 99.1   \\
  \hline
  
\end{tabular}
\caption{\label{results}
Stage 1 Monte Carlo: Fine-grained experimental results for {\color{BrickRed} \textbf{Extractive}},
{\color{ForestGreen} \textbf{Multiple Choice}}, and {\color{Mulberry} \textbf{Abstractive}} question-answering scenarios. We display each dataset, split, and corresponding metrics for accuracy, ensemble accuracy (Mode), lower \& upper bounds for accuracy, item difficulty ($\mathbf{\mu_D}$), agreement (Fleiss's Generalized $\mathbf{\kappa}$, Mean Certainty $\mathbf{H_{\eta}}$, Gibb's $\mathbf{M_2}$ Index), and reliability (Cronbach's $\mathbf{\alpha}$). For MathQA train we only evaluated 693, out of a possible 29,800 samples.
}
\end{table*}

\begin{table*}[!th]
\centering
\begin{tabular}{cll|cc|cc|ccccccc}
\hline
\multicolumn{3}{c|}{} & \multicolumn{2}{c|}{\textbf{Hallucination}} & \multicolumn{2}{c|}{\textbf{At least one}} & \multicolumn{7}{c}{\textbf{Percent of}} \\
\multicolumn{3}{c|}{\textbf{Datasets}} & \multicolumn{2}{c|}{\textbf{Rate}} & \multicolumn{2}{c|}{\textbf{hallucination?}} & \multicolumn{7}{c}{\textbf{Hallucinated Agents per Query} }  \\
\hline
& \textbf{Name} & \textbf{Split} & \multicolumn{1}{c}{\textbf{No}} & \multicolumn{1}{c|}{\textbf{Yes}} & \multicolumn{1}{c}{\textbf{No}} & \multicolumn{1}{c|}{\textbf{Yes}} & \multicolumn{1}{c}{\textbf{0}} & \multicolumn{1}{c}{\textbf{1}} & \multicolumn{1}{c}{\textbf{2}} & \multicolumn{1}{c}{\textbf{3}} & \multicolumn{1}{c}{\textbf{4}} & \multicolumn{1}{c}{\textbf{5}} & \multicolumn{1}{c}{\textbf{6}} \\
\hline\hline
\multirow{3}{*}{\rotatebox[origin=c]{90}{{\color{BrickRed} \textbf{Extn}}}} &
 \multirow{2}{*}{SQuADv2} & train &  87.0 & 13.0 & 68.6 &  31.4 & 68.6 & 13.0 & 6.2 & 4.1 & 2.9 & 2.6 & 2.7 \\
     &    & val & 90.6 & 9.4 & 74.5 & 25.5 & 74.5 & 12.1 & 5.4 & 2.9 & 2.1 & 1.8 & 1.2 \\
\cline{2-14}
 & {\color{BrickRed}\textbf{Total}} & -  & 87.2 & 12.8 & 69.0 & 31.0 & 69.0 & 12.9 & 6.1 & 4.0 & 2.9 & 2.5 & 2.6 \\
\hline\hline
 \multirow{24}{*}{\rotatebox[origin=c]{90}{{\color{ForestGreen} \textbf{Multiple Choice}}}} &
 TruthfulQA & val  & 58.5 & 41.5 & 39.8 & 60.2 & 39.8 & 10.3 & 6.1 & 5.1 & 6.4 & 8.4 & 23.9 \\
 \cline{2-14}
 &\multirow{2}{*}{PIQA} & train & 78.8 & 21.2 & 56.7 & 43.3 & 56.7 & 13.7 & 7.7 & 6.0 & 5.2 & 4.7 & 6.0  \\
 &     & val & 79.2 & 20.8 & 58.2 & 41.8 & 58.2 & 11.5 & 8.9 & 6.7 & 4.1 & 4.3 & 6.2  \\
\cline{2-14}
 & \multirow{3}{*}{MMLU} & dev & 58.9 & 41.1 & 35.6 & 64.4 & 35.6 & 13.2 & 5.7 & 6.0 & 11.0 & 11.0 & 17.4  \\
&  & val & 60.3 & 39.7 & 37.9 & 62.1 & 37.9 & 11.3 & 7.5 & 6.6 & 8.2 & 11.3 & 17.2  \\
&  & test & 61.6 & 38.4 & 38.4 & 61.6 & 38.4 & 11.3 & 8.1 & 7.3 & 8.9 & 10.2 & 15.8  \\
 \cline{2-14}
 & \multirow{3}{*}{\begin{tabular}{@{}l} OpenBook \\ QA \end{tabular}} & train &   67.8 & 32.2 & 37.8 & 62.2 & 37.8 & 16.9 & 11.2 & 8.7 & 8.0 & 8.7 & 8.7 \\
 & & val &  70.5 & 29.5 & 39.8 & 60.2 & 39.8 & 18.5 & 9.9 & 9.1 & 9.1 & 6.6 & 7.0 \\
 & & test &  66.8 & 33.2 & 38.1 & 61.9 & 38.1 & 16.0 & 9.6 & 8.4 & 9.6 & 9.0 & 9.2 \\
 \cline{2-14}
 & \multirow{2}{*}{BoolQ} & train & 67.0 & 33.0 & 32.7 & 67.3 & 32.7 & 18.5 & 13.9 & 10.6 & 9.0 & 7.9 & 7.4  \\
 & & val &  67.6 & 32.4 & 33.5 & 66.5 & 33.5 & 18.7 &13.2 & 10.7 & 9.0 & 8.0 & 6.8 \\
 \cline{2-14}
 &\multirow{3}{*}{SciQ} & train &  89.9 & 10.1 & 76.7 & 23.3 & 76.7 & 9.3 & 4.3 & 2.9 & 2.3 & 2.1 & 2.4\\
 & & val &  89.9 & 10.1 & 77.3 & 22.7 & 77.3 & 8.4 & 5.0 & 2.6 & 1.7 & 2.3 & 2.7 \\
 & & test & 90.6 & 9.4 & 76.9 & 23.1 & 76.9 & 11.0 & 4.0 & 1.8 & 2.0 & 2.1 & 2.2  \\
 \cline{2-14}
 &\multirow{3}{*}{\begin{tabular}{@{}l} ARC - \\ Challenge \end{tabular}} & train &  76.8 & 23.2 & 52.3 & 47.7 & 52.3 & 14.4 & 8.9 & 6.8 & 6.4 & 6.4 & 4.8 \\
 & & val & 77.1 & 22.9 & 53.8 & 46.2 & 53.8 & 11.7 & 10.0 & 6.7 & 8.0 & 4.7 & 5.0 \\
 & & test & 74.7 & 25.3 & 50.9 & 49.1 & 50.9 & 14.2 & 8.0 & 7.2 & 6.1 & 6.2 & 7.3 \\
 \cline{2-14}
 &\multirow{3}{*}{\begin{tabular}{@{}l} ARC - \\ Easy \end{tabular}} & train  & 88.0 & 12.0 & 70.8 & 29.2 & 70.8 & 12.0 & 6.0 & 3.5 & 3.2 & 2.4 & 2.1  \\
 & & val  &  86.7 & 13.3 & 66.1 & 33.9 & 66.1 & 16.3 & 5.3 & 3.9 & 2.5 & 4.0 & 1.9  \\
 & & test  &  87.9 & 12.1 & 69.6 & 30.4 & 69.6 & 13.6 & 5.7 & 3.3 & 3.2 & 2.9 & 1.7 \\
 \cline{2-14}
 &\multirow{3}{*}{MathQA} & train  & 46.6 & 53.4 & 9.5 & 90.5 & 9.5 & 12.8 & 14.7 & 16.6 & 17.2 & 15.0 & 14.1 \\
 & & val  &  45.9 & 54.1 & 9.4 & 90.6 & 9.4 & 12.5 & 14.9 & 15.9 & 16.3 & 16.9 & 14.2  \\
 & & test & 45.7 & 54.3 & 9.6 & 90.4 & 9.6 & 13.0 & 14.1 & 15.3 & 16.2 & 16.1 & 15.5 \\
 \cline{2-14}
 & {\color{ForestGreen}\textbf{Total}} & -  & 71.8 & 28.2 & 47.4 & 52.6 & 47.4 & 13.3 & 9.0 & 7.5 & 7.2 & 7.2 & 8.4 \\
\hline\hline
\multirow{14}{*}{\rotatebox[origin=c]{90}{{\color{Mulberry} \textbf{Abstractive}}}}
 & \multirow{2}{*}{SQuADv2} & train  &  29.9 & 70.1 & 15.1 & 84.9 & 15.1 & 6.3 & 5.1 & 5.3 & 5.9 & 9.0 & 53.3 \\
   &      & val  &  22.9 & 77.1 & 10.2 & 89.8 & 10.2 & 5.5 & 4.3 & 3.9 & 5.5 & 8.5 & 62.1 \\
 \cline{2-14}
 & TruthfulQA & val &  55.1 & 44.9 & 28.1 & 71.9 & 28.1 & 12.9 & 10.9 & 9.9 & 7.2 & 9.5 & 21.4  \\
 \cline{2-14}
 & \multirow{3}{*}{WikiQA} & train  & 69.8 & 30.2 & 54.6 & 45.4 & 54.6 & 10.4 & 4.5 & 3.1 & 3.5 & 4.9 & 19.1 \\
 & & val  & 73.3 & 26.7 & 58.6 & 41.4 & 58.6 & 10.0 & 5.0 & 2.9 & 3.6 & 2.9 & 17.1 \\
 & & test  &  67.8 & 32.2 & 52.4 & 47.6 & 52.4 & 7.7 & 6.3 & 4.2 & 5.2 & 5.2 & 18.9  \\

 \cline{2-14}
 & \multirow{3}{*}{SciQ} & train  & 59.8 & 40.2 & 35.0 & 65.0 & 35.0 & 13.2 & 9.1 & 7.7 & 7.3 & 8.0 & 19.6 \\
 & & val  & 59.0 & 41.0 & 35.2 & 64.8 & 35.2 & 11.8 & 9.4 & 7.5 & 7.3 & 9.3 & 19.6  \\
 & & test  &  61.9 & 38.1 & 35.5 & 64.5 & 35.5 & 13.7 & 11.6 & 8.0 & 5.4 & 9.0 & 16.8 \\
 \cline{2-14}
 & \multirow{2}{*}{\begin{tabular}{@{}l} HotpotQA \\ (KILT) \end{tabular}} & train  &  40.6 & 59.4 & 21.2 & 78.8 & 21.2 & 9.7 & 6.9 & 6.1 & 6.6 & 9.1 & 40.4 \\
 & & val &  38.3 & 61.7 & 21.5 & 78.5 & 21.5 & 8.0 & 6.0 & 5.3 & 6.4 & 8.9 & 44.1 \\
 \cline{2-14}
& \multirow{2}{*}{TriviaQA} & train  & 66.8 & 33.2 & 48.6 & 51.4 & 48.6 & 11.7 & 6.0 & 4.5 & 3.9 & 4.9 & 20.3 \\
 & & dev  &  67.0 & 33.0 & 48.5 & 51.5 & 48.5 & 12.0 & 6.4 & 4.3 & 3.9 & 5.0 & 19.9  \\
 \cline{2-14}
 & {\color{Mulberry}\textbf{Total}} & - & 51.9 & 48.1 & 33.4 & 66.6 & 33.4 & 10.3 & 6.4 & 5.3 & 5.4 & 7.2 & 32.1 \\
  \hline
  
\end{tabular}
\caption{\label{hallucination_rates}
Hallucination rates per dataset. The first two columns report the individual agent hallucination rate in total. For example, 87.0\% of agents responded correctly for SQuADv2, while 13.0\% hallucinated. This metric does not translate into our binary or expected values, as the latter are aggregated on a sample basis, while hallucination rates are global measures. Next, we report the binary label breakdown per split, along with the expected value labels.
}
\end{table*}


\begin{table*}[!ht]
\centering
\begin{tabular}{cllr|rr|rr|rr|rr}
\hline
\multicolumn{4}{c|}{\textbf{Datasets}} & \multicolumn{2}{c|}{\textbf{Consensus} $\mathbf{\uparrow}$} & \multicolumn{2}{c|}{\textbf{Dissent} $\mathbf{\downarrow}$} & \multicolumn{2}{c|}{\textbf{Corrective} $\mathbf{\uparrow}$}  & \multicolumn{2}{c}{\textbf{Erroneous} $\mathbf{\downarrow}$} \\
\hline
& \textbf{Name} & \textbf{Split} & \textbf{\#} & \textbf{\#} & \textbf{\%} & \textbf{\#} & \textbf{\%} & \textbf{\#} & \textbf{\%} & \textbf{\#} & \textbf{\%} \\
\hline\hline
\multirow{3}{*}{\rotatebox[origin=c]{90}{{\color{BrickRed} \textbf{Extn}}}} &
 \multirow{2}{*}{SQuADv2} & train & 80,049 &   71,560 & 89.4 & 2,037 & 2.5 & 1,121 & 1.4 & 5,331 & 6.7 \\
     &    & val & 5,843 & 5,440 & 93.1 & 120 & 2.1 & 58 & 1.0 & 225 & 3.9\\
     \cline{2-12}
     & {\color{BrickRed} \textbf{Total}} & - & 85,892 & 77,000 & 89.6 & 2,157 & 2.5 & 1,1790 & 1.4 & 5,556 & 6.5 \\
\hline\hline
 \multirow{24}{*}{\rotatebox[origin=c]{90}{{\color{ForestGreen} \textbf{Multiple Choice}}}} &
 TruthfulQA & val & 786 & 443 & 56.4 & 32 & 4.1 & 31 & 3.9 & 280 & 35.6  \\
 \cline{2-12}
 &\multirow{2}{*}{PIQA} & train & 15,677 &  12,234 & 78.0 & 494 & 3.2 & 675 & 4.3 & 2,274 & 14.5 \\
 &     & val & 1,784 & 1,403 & 78.6 & 41 & 2.3 & 82 & 4.6 & 258 & 14.5  \\
\cline{2-12}
 & \multirow{3}{*}{MMLU} & dev & 281 & 164 & 58.4 & 22 & 7.8 & 10 & 3.6 & 85 & 30.2  \\
&  & val & 1,463 & 875 & 59.8 & 68 & 4.6 & 76 & 5.2 & 444 & 30.3  \\
&  & test & 13,545 &  8,415 & 62.1 & 746 & 5.5 & 696 & 5.1 & 3,688 & 27.2 \\
 \cline{2-12}
 & \multirow{3}{*}{\begin{tabular}{@{}l} OpenBook \\ QA \end{tabular}} & train & 4,909 &  3,475 & 70.8 & 354 & 7.2 & 238 & 4.8 & 842 & 17.2 \\
 & & val & 497 & 341 & 68.3 & 36 & 7.2 & 27 & 5.4 & 95 & 19.0  \\
 & & test & 499 & 361 & 72.6 & 27 & 5.4 & 31 & 6.2 & 78 & 15.7  \\
 \cline{2-12}
 & \multirow{2}{*}{BoolQ} & train & 9,401 & 6,238 & 66.4 & 438 & 4.7 & 458 & 4.9 & 2,267 & 24.1  \\
 & & val & 3,256 & 2,180 & 67.0 & 147 & 4.5 & 160 & 4.9 & 769 & 23.6  \\
 \cline{2-12}
 &\multirow{3}{*}{SciQ} & train & 11,670 & 10,690 & 91.6 & 205 & 1.8 & 152 & 1.3 & 623 & 5.3\\
 & & val & 999 &  903 & 90.4 & 12 & 1.2 & 26 & 2.6 & 58 & 5.8 \\
 & & test & 998 & 921 & 92.3 & 15 & 1.5 & 12 & 1.2 & 50 & 5.0  \\
 \cline{2-12}
 &\multirow{3}{*}{\begin{tabular}{@{}l} ARC - \\ Challenge \end{tabular}} & train & 1,118 & 899 & 80.4 & 57 & 5.1 & 28 & 2.5 & 134 & 12.0  \\
 & & val & 299 & 241 & 80.6 & 19 & 6.4 & 5 & 1.7 & 34 & 11.4 \\
 & & test & 1,172 & 913 & 77.9 & 69 & 5.9 & 33 & 2.8 & 157 & 13.4 \\
 \cline{2-12}
 &\multirow{3}{*}{\begin{tabular}{@{}l} ARC - \\ Easy \end{tabular}} & train & 2,248 & 2,040 & 90.7 & 58 & 2.6 & 43 & 1.9 & 107 & 4.8   \\
 & & val & 570 & 517 & 90.7 & 21 & 3.7 & 8 & 1.4 & 24 & 4.2   \\
 & & test & 2,374 & 2,149 & 90.5 & 53 & 2.2 & 46 & 1.9 & 126 & 5.3  \\
 \cline{2-12}
 &\multirow{3}{*}{MathQA} & train* & 693 & 306 & 44.2 & 41 & 5.9 & 88 & 12.7 & 258 & 37.2  \\
 & & val & 4,473 &  1,902 & 42.5 & 325 & 7.3 & 582 & 13.0 & 1,664 & 37.2  \\
 & & test & 2,985 & 1,252 & 41.9 & 171 & 5.7 & 382 & 12.8 & 1,180 & 39.5 \\
 \cline{2-12}
     & {\color{ForestGreen} \textbf{Total}} & - & 81,697 & 58,862 & 72.0 & 3,451 & 4.2 & 3,889 & 4.8 & 15,495 & 19.0 \\
\hline\hline
\multirow{14}{*}{\rotatebox[origin=c]{90}{{\color{Mulberry} \textbf{Abstractive}}}}
 & \multirow{2}{*}{SQuADv2} & train & 20,842 & 5,884 & 28.2 & 980 & 4.7 & 352 & 1.7 & 13,626 & 65.4  \\
   &      & val & 5,864 & 1,236 & 21.1 & 254 & 4.3 & 79 & 1.3 & 4,295 & 73.2  \\
 \cline{2-12}
 & TruthfulQA & val & 807 & 397 & 49.2 & 26 & 3.2 & 67 & 8.3 & 317 & 39.3   \\
 \cline{2-12}
 & \multirow{3}{*}{WikiQA} & train & 1,028 & 720 & 70.0 & 35 & 3.4 & 15 & 1.5 & 258 & 25.1 \\
 & & val & 140 & 104 & 74.3 & 3 & 2.1 & 2 & 1.4 & 31 & 22.1 \\
 & & test & 286 &  194 & 67.8 & 17 & 5.9 & 6 & 2.1 & 69 & 24.1  \\
 \cline{2-12}
 & \multirow{3}{*}{SciQ} & train & 11,596 &  7,035 & 60.7 & 660 & 5.7 & 371 & 3.2 & 3,530 & 30.4 \\
 & & val & 991 & 585 & 59.0 & 62 & 6.3 & 32 & 3.2 & 312 & 31.5  \\
 & & test & 995 & 634 & 63.7 & 58 & 5.8 & 27 & 2.7 & 276 & 27.7  \\
 \cline{2-12}
 & \multirow{2}{*}{\begin{tabular}{@{}l} HotpotQA \\ (KILT) \end{tabular}} & train & 66,345 & 26,207 & 39.5 & 3,968 & 6.0 & 1,495 & 2.3 & 34,675 & 52.3  \\
 & & val & 5,542 & 2,037 & 36.8 & 318 & 5.7 & 117 & 2.1 & 3,074 & 55.4  \\
 \cline{2-12}
& \multirow{2}{*}{TriviaQA} & train & 76,635 & 52,051 & 67.9 & 2,929 & 3.8 & 1,188 & 1.6 & 20,467 & 26.7  \\
 & & dev & 11,177 & 7,646 & 68.4 & 427 & 3.8 & 164 & 1.5 & 2,940 & 26.3    \\
 \cline{2-12} 
 & {\color{Mulberry} \textbf{Total}} & - & 202,248 & 104,713 & 51.8 & 9,735 & 4.8 & 3,921 & 1.9 & 83,879 & 41.5 \\
  \hline\hline
  & - & train & 302,492 & 199,487 & 65.9 & 12,271 & 4.1 & 6,231 & 2.1 & 84,503 & 27.9 \\
  & - & val   & 44,491  & 26,270 & 59.0 & 1,905 & 4.3 & 1,525 & 3.4 & 14,791 & 33.2 \\
  & - & test  & 22,854 & 14,818 & 64.8 & 1,167 & 5.1 & 1,233 & 5.4 & 5,636 & 24.7 \\
  \cline{2-12}
  &  Average & - & - & - & 66.9 & - & 4.5 & - & 3.8 & - & 24.8
  \\
  &  Dispersion & - & - & - & 18.9 & - & 1.8 & - & 3.2 & - & 17.0
  \\ 
  &  \textbf{Grand Total} & - & 369,837 & 240,575 & 65.0 & 15,343 & 4.1 & 8,989 & 2.4 & 104,930 & 28.4
  \\
  \hline
  
\end{tabular}
\caption{\label{consensus_results}
We investigate the four main scenarios encountered during Stage 1 aggregation. \textbf{Consensus} arises when the original query and the majority of perturbed queries produce the correct answer. \textbf{Dissent} arises when the majority of agents disagree with the correct output generated for the original query. \textbf{Corrective} arises when the original query's generated answer is incorrect, but the majority agrees on the correct answer. Finally, \textbf{Erroneous} arises when both the original query's answer and the majority of agents are incorrect. The Consensus and Corrective scenarios contribute to improvements in accuracy, while the Dissent and Erroneous cases result in lower performance. As seen from the table, \texttt{gpt-3.5-turbo} is consistent with the majority of perturbations - regardless of correctness. This indicates that, under perturbations, \texttt{gpt-3.5-turbo} can make new mistakes.
}
\end{table*}

\begin{figure*}[t]
\begin{lstlisting}[caption={Our PyTorch implementation of an Ordinal Layer, which accepts scalar values from any model and outputs a multi-class probability distribution for \texttt{n\_classes}.},captionpos=b]
# Ordinal.py
import torch
import torch.nn as nn
import torch.nn.functional as F

# Base Layer For Ordinal Prediction
class OrdinalLayer(nn.Module):
    def __init__(self, n_classes, func=torch.sigmoid):
        super().__init__()
        self.func = func
        self.theta = nn.Parameter(torch.linspace(-1, 1, n_classes - 1))
        self.mask = torch.tensor([1] + [0 for _ in range(n_classes - 2)])

    def forward(self, x, return_prob=False):
        # B: Batch Size
        # Input: x -> (B, *, 1)
        size = x.size()
        x = self.threshold - x.view(-1, 1)

        x = torch.cat((
                torch.zeros(x.size(0), 1),
                self.func(x), # any cdf
                torch.ones(x.size(0), 1)
            ), dim=-1)

        x = x[:, 1:] - x[:, :-1]

        # Directly gives log probs,
        # Use NLL as they can not be softmaxed
        # Return: Log Probs
        # x -> (B, *, N_CLASSES)
        if return_prob:
            return x.view(*size[:-1], -1)
        return (x + 1e-8).log().view(*size[:-1], -1)

    @property
    def threshold(self):
        return (self.theta * self.mask + F.softplus(self.theta) * (1 - self.mask)).cumsum(-1)


# Wrapped Loss to avoid Softmax
class OrdinalLoss(nn.Module):
    def __init__(self, **kwargs):
        super().__init__()
        self.loss = nn.NLLLoss(**kwargs)

    def forward(self, x, y):
        # x -> Logits, size: (B, C)
        # y -> Labels, list (B) (like cross entropy)
        return self.loss(x, y)
\end{lstlisting}
\end{figure*}

\end{document}